\documentclass[10pt,twocolumn,letterpaper]{article}

\usepackage{cvpr}              
\usepackage{multirow}
\usepackage{graphicx}
\usepackage{subcaption} 
\usepackage[accsupp]{axessibility}  

\definecolor{cvprblue}{rgb}{0.21,0.49,0.74}
\usepackage[pagebackref,breaklinks,colorlinks,allcolors=cvprblue]{hyperref}

\def\paperID{} 
\def\confName{CVPR}
\def\confYear{2026}

\title{HOLO: Homography-Guided Pose Estimator Network for Fine-Grained Visual Localization on SD Maps}

\author{Xuchang Zhong$^1$ \quad 
Xu Cao$^1$ \quad
Jinke Feng$^2$ \quad
Hao Fang$^{1,*}$\\
$^1$Beijing Institute of Technology \quad 
$^2$University of Science and Technology of China\\
{\tt\small \{xuchang.zhong, xucao, fangh\}@bit.edu.cn \quad fengjinke@mail.ustc.edu.cn}
}

\begin{document}
\maketitle
\begin{abstract}
Visual localization on standard-definition (SD) maps has emerged as a promising low-cost and scalable solution for autonomous driving. However, existing regression-based approaches often overlook inherent geometric priors, resulting in suboptimal training efficiency and limited localization accuracy. In this paper, we propose a novel homography-guided pose estimator network for fine-grained visual localization between multi-view images and standard-definition (SD) maps. We construct input pairs that satisfy a homography constraint by projecting ground-view features into the BEV domain and enforcing semantic alignment with map features. Then we leverage homography relationships to guide feature fusion and restrict the pose outputs to a valid feasible region, which significantly improves training efficiency and localization accuracy compared to prior methods relying on attention-based fusion and direct 3-DoF pose regression. To the best of our knowledge, this is the first work to unify BEV semantic reasoning with homography learning for image-to-map localization. Furthermore, by explicitly modeling homography transformations, the proposed framework naturally supports cross-resolution inputs, enhancing model flexibility. Extensive experiments on the nuScenes dataset demonstrate that our approach significantly outperforms existing state-of-the-art visual localization methods. Code and pretrained models will be publicly available at \href{https://github.com/Quartararo0714/HOLO/}{HOLO}.
\end{abstract}    
\section{Introduction}
\label{sec:intro}

Reliable ego localization is a fundamental requirement for autonomous driving but remains challenging in complex and dynamic environments. GPS-based localization suffers from severe signal degradation in urban canyons, tunnels, and occluded regions, leading to large positioning drift. Visual localization, which estimates vehicle pose from camera observations, offers a promising solution in such GPS-denied scenarios.

\begin{figure}[t]
  \centering
   \includegraphics[width=0.97\linewidth]{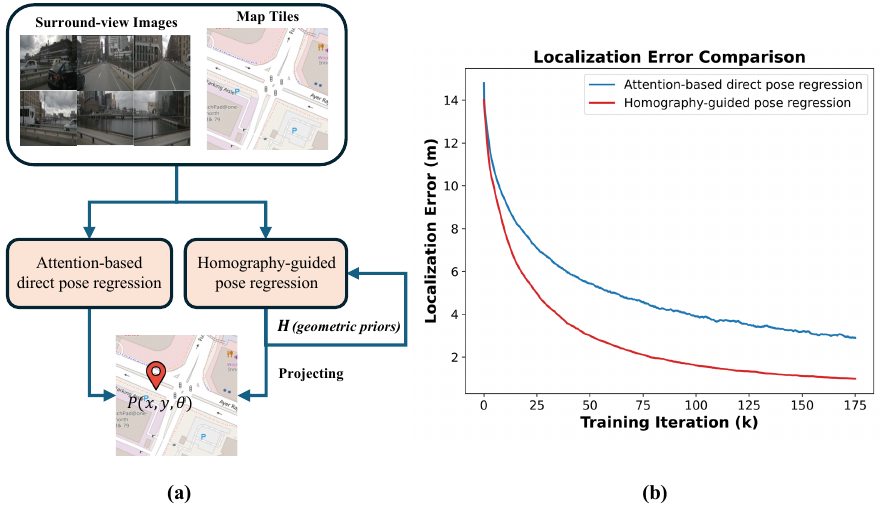}
   \caption{Comparision of attention-based direct pose estimation and homography-guided pose estimation. (a) Schematic diagram of attention-based direct pose estimation and homography-guided pose estimation. (b) The evolution of localization error across training iterations between two methods on nuScene\cite{caesar2020nuscenes}.}
   \label{intro}
\end{figure}

Recent studies have explored diverse map representations for localization, including satellite maps \cite{lentsch2023slicematch,wang2023fine,xia2023convolutional,yuan2024cross}, high-definition (HD) maps \cite{he2024egovm,zhang2025bev,Zhao2024towards}, and 3D prior maps \cite{chen2024map,Dong_2025_CVPR}.
In autonomous driving scenarios, early research on visual localization mainly relied on HD maps. However, HD maps are costly to construct, require frequent maintenance, and lack scalability. These limitations have motivated growing interest in using lightweight and easily accessible standard-definition maps (SD) maps, such as OpenStreetMap \cite{Haklay2008open}, for large-scale localization.

Early works \cite{samano2020you, zhou2021efficient} formulated visual localization as an image retrieval task to perform coarse localization.
With the rise of bird’s-eye-view (BEV) perception\cite{philion2020lift,li2024bevformer,li2023fb}, research has shifted toward geometric alignment between BEV features and map priors for improved accuracy. 
OrienterNet \cite{sarlin2023orienternet} introduced an end-to-end framework performing exhaustive pose-wise BEV-to-map matching but suffered from high computational cost. 
To enhance efficiency, MapLocNet \cite{wu2024maplocnet} employs a regression-based approach to directly predict the target pose, avoiding the complex matching process. However, while matching methods only need to distinguish similarities within a limited set of pose candidates, regression requires fitting continuous pose values, which makes training more challenging and convergence slower.

To address this problem, we identify two main limitations in existing regression-based approaches. First, during feature fusion, the absence of explicit geometric guidance leads most methods \cite{wu2024maplocnet, he2024egovm, zhang2025bev} to rely solely on attention-based mechanisms to learn cross-modal correlations, which often results in low efficiency. Second, directly regressing a 3-DoF pose without geometric constraints introduces unstable gradients, making optimization challenging and prone to overfitting.

To overcome these issues, we observe that a local BEV representation and the corresponding map tile inherently exhibit a homography. Thus, we propose a novel homography-guided multi-view pose estimator network with coupled semantic alignment, namely HOLO. The key idea is to exploit homography-based geometric priors between BEV features and map representations to provide explicit spatial correspondence and geometric guidance for pose estimation. Specifically, we first encode multi-view surround images into the BEV space and enforce semantic consistency with rasterized SD maps to form homography pairs. Instead of relying purely on attention for feature correspondence, the learned homography relations are dynamically fed back into the network to guide cross-feature correlation computation. Furthermore, to impose geometric constraints on the output, we first predict corner displacements to derive the homography matrix and then recover the 3-DoF pose through geometric mapping. Our framework jointly optimizes semantic alignment and weakly supervised homography estimation in an end-to-end fashion, where the two tasks reinforce each other—yielding faster convergence and higher localization accuracy.

To promote further research on visual localization with SD maps, we collected OpenStreetMap (OSM) data corresponding to the four cities in the nuScenes\cite{caesar2020nuscenes} dataset, which we then aligned with the provided HD maps and plan to release publicly. Experimental results demonstrate that our approach significantly outperforms the current state-of-the-art visual localization method. The main contributions of this work are summarized as follows:
\begin{itemize}
    \item[1)] We propose a novel framework for multi-camera fine-grained visual localization by reformulating the task as a homography estimation problem between BEV representations and SD maps. We incorporate the geometric relationships between the two modalities into both the feature fusion and pose decoding modules, which improves convergence speed and localization accuracy.
    \item[2)] Our framework unifies BEV semantic reasoning and homography estimation into a single training pipeline, enabling homography learning directly from semantic features. The joint optimization also improves both localization accuracy and semantic representation quality.
    \item[3)] Our model achieves state-of-the-art accuracy for visual localization on the nuScenes dataset, improving Recall@1m/2m by 16\% and achieving faster convergence. Extensive experiments demonstrate the robustness and efficiency of the proposed framework, which runs at 20 FPS during inference.
\end{itemize}


\begin{figure*}
  \centering
  \includegraphics[width=0.95\linewidth]{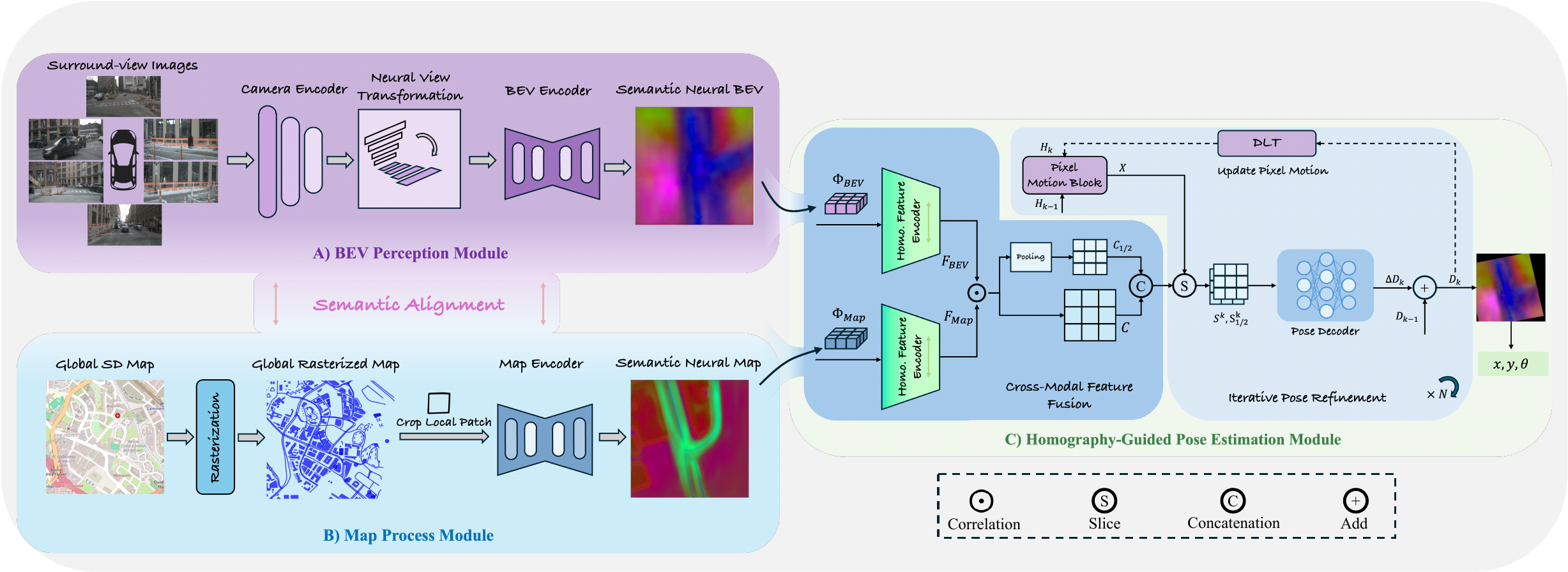}
  \caption{Overall architecture of the proposed Homography-Guided Pose Estimator Network. The BEV Perception Module and Map Processing Module collaboratively build neural feature pairs with homography through semantic alignment, providing explicit geometric priors for downstream pose regression. The Homography-Guided Pose Estimation Module leverages homography priors to accomplish the final pose estimation.}
  \label{fig:pipeline}
\end{figure*}

\section{Related Work}
\label{sec:related}
\subsection{Visual Localization with Maps}

Visual localization on maps is a fundamental task in autonomous systems. Although the implementation of visual localization may differ across tasks depending on the type of map representation, the underlying principles of pose estimation remain consistent. Image-to-SD map localization mainly consists of two technical paradigms: matching-based methods and regression-based methods.

\textbf{Matching-based methods.}
These methods formulate image-to-map localization as a pose retrieval problem, where the goal is to learn discriminative embeddings for both images and maps and maximize their similarity at the corresponding poses. Some works \cite{lentsch2023slicematch,xia2023convolutional} focus on encoding ground-view and aerial-view images into pose-aware feature vectors that jointly capture spatial and orientation information for retrieval, while others \cite{sarlin2023orienternet,sarlin2023snap} extend this idea by learning implicit BEV features and projecting them onto the map according to pose candidates to compute matching scores, where the highest-scoring hypothesis is selected as the final pose. Based on this framework, SegLocNet\cite{zhou2025seglocnet} further replaces implicit features with binarized semantic masks to improve localization precision.
Although these methods achieve strong performance through representation learning, they exhibit inherent limitations.
Since they rely on discrete pose sampling for feature matching, their localization accuracy is constrained by the sampling resolution.
Moreover, as the number of pose hypotheses increases, the inference time grows significantly, resulting in a trade-off between accuracy and efficiency.

\textbf{Regression-based methods.} 
Unlike matching-based approaches that require pose discretization, regression-based methods infer the camera pose directly from paired image and map inputs.
These approaches generally comprise three core components: feature extraction, cross-modal fusion, and pose decoding. Recent approaches \cite{chen2024map,wu2024maplocnet} leverage transformer-based encoders for effective feature fusion, whereas subsequent studies \cite{Dong_2025_CVPR, yuan2024cross, zhang2025bev} further enhance inter-modal interactions through cross-attention mechanisms.
Most of these methods focus on designing robust fusion modules to obtain features that can be more easily mapped to pose, thereby improving localization accuracy. However, they mainly relied on attention mechanisms to learn feature associations implicitly, without incorporating any geometric priors into this stage. This often led to low efficiency and poor accuracy.

\subsection{Cross-modal Homography Estimation}

Homography estimation aims to recover a planar projective transformation between two views by estimating a 3x3 homography matrix that maps corresponding points across image planes. In cross-modal settings, the two inputs originate from different sensing modalities or data sources, such as RGB vs. NIR or optical vs. SAR imagery. Early deep homography methods \cite{zhao2021deep,cao2022iterative,cao2023recurrent} extended supervised homography learning to the cross-modal case under strong direct supervision. However, ground-truth homography annotations are rarely available in real-world scenarios, which significantly limits the practicality of such approaches. 

To alleviate this constraint, recent works explore unsupervised cross-modal homography estimation. Zhang \etal \cite{zhang2024scpnet} align heterogeneous image pairs through a consistent feature projection module and leverage intra-modal self-supervision to compensate for missing ground-truth labels. Song \etal \cite{song2024unsupervised} alternate between modality adaptation and geometric refinement to progressively reduce appearance and alignment discrepancies. Yu \etal \cite{yu2025sshnet} further decomposes the problem into modality alignment and homography regression, optimizing the two sub-networks separately to stabilize training. 

Inspired by this decomposition strategy, we formulate multi-view image to SD map localization in a similar manner: we first reduce the modality gap between images and SD maps using a semantic alignment network, and then regress the relative pose using a homography estimation network. Unlike \cite{yu2025sshnet}, which requires stage-wise training due to the absence of geometric supervision, our task benefits from available GT vehicle poses, allowing us to jointly optimize both components in an end-to-end manner.


\section{Methodology}
\label{sec: method}

The architecture of our proposed Homography-Guided Pose Estimator Network is illustrated in \cref{fig:pipeline}.  Our network mainly consists of a BEV perception module, a map processing module, and a homography-guided pose estimation module. The entire network is jointly optimized in an end-to-end manner using both semantic and pose losses.

\subsection{Problem Formulation}
\label{subsec: PF}

Given a set of multi-view images $\mathcal{I} = \{ I_i \}_{i=1}^{N}$ captured by on-board cameras and a reference map $\mathcal{U}$ determined by noisy GPS signals, the goal of our system is to estimate the accurate vehicle's pose on the map. Since the map is defined on a two-dimensional plane, the localization task can be simplified into a 3-degree-of-freedom (3-DoF) pose estimation problem. Specifically, the vehicle pose can be represented as
\begin{equation}
    \mathbf{p} = (x, y, \theta),
\end{equation}
where $(x, y) \in \mathbb{R}^2$ denotes the vehicle's position on the map, and $\theta \in \left[ -\pi,\pi \right]$ represents its heading angle around the vertical ($z$-) axis. The coordinate system follows the East-North-Up (ENU) convention.

The goal of the system is to learn a function
\begin{equation}
    f_{\boldsymbol{\theta}}: (\mathcal{I}, \mathcal{U}) \rightarrow \mathbf{p},
\end{equation}
parameterized by network weights $\boldsymbol{\theta}$, that predicts the vehicle pose $\mathbf{p}$ aligning the camera observations with the map domain.

\subsection{BEV Perception Module}
\label{subsec: bev}

Our BEV perception module aims to project the surround-view images \(I\) into the BEV space to reduce the perspective gap between camera observations and the map for downstream homography estimation tasks. 
Following prior works such as LSS \cite{philion2020lift} and OrienterNet \cite{sarlin2023orienternet}, we adopt a depth-guided projection mechanism that transforms image features into BEV representations in a simple yet effective manner. 

\textbf{Camera Feature Extraction:} We utilized a lightweight pre-trained EfficientNet-B0\cite{tan2019efficientnet} backbone to extract multi-scale semantic features from all six ground images. To model scene geometry, the network jointly predicts a discrete depth distribution and semantic embeddings through a shared head, generating depth-aware volumetric features.

\textbf{Neural View Transformation:}
Using known camera intrinsics \(\mathbf{K}\) and extrinsics \((\mathbf{R}, \mathbf{t})\), the volumetric features are lifted into the 3D ego-vehicle coordinate frame and aggregated via differentiable voxel pooling to form a dense BEV tensor \(\mathbf{B} \in \mathbb{R}^{C_d \times H_{bev} \times W_{bev}}\).

\textbf{BEV Feature Encoding:}
The BEV tensor \(\mathbf{B}\) is further encoded by a BEV encoder to enhance semantic abstraction, producing the final BEV representation $\mathbf{\Phi}_{bev} \in \mathbb{R}^{C \times H_{bev} \times W_{bev}} $. Then, in order to improve the consistency with map features, a segmentation head generates task-specific outputs \(\mathbf{M}_{bev}^{sem} \in \mathbb{R}^{2 \times H_{bev} \times W_{bev}}\) for semantic alignment learning.

\subsection{Map Process Module}
\label{subsec: map}

We design a map processing module that transforms the SD map into a feature representation compatible with the BEV feature space.

\textbf{Map Rasterization:} We adopt OpenStreetMap as the input source for our map processing module.
However, the raw OSM data encodes map elements as vector-based geometric primitives (\eg polygons, lines, and points) with diverse semantic categories, which can't be directly processed by neural networks. 
To this end, we first perform rasterization to convert vector-based map elements into a grid-based representation. 
Considering the limited spatial accuracy and inconsistent updates of OSM maps, which may lead to discrepancies with camera observations, we discard minor elements and retain only the two most critical classes: roads and buildings.
As a result, we obtain a two-channel local map patch $\mathbf{U} \in \mathbb{R}^{2 \times H_{\text{map}} \times W_{\text{map}}}$, which is then fed into our map process network for feature extraction and alignment.

\textbf{Map Encoding:} Similar to OrienterNet \cite{sarlin2023orienternet}, we first assign an $N$-dimensional learnable embedding to each class in the rasterized map, producing a feature map of size $H \times W \times 2N$.
The embedded map is then processed by a U-Net architecture built upon VGG16 \cite{vgg} to extract semantic features.
The resulting map feature $\mathbf{\Phi}_{map} \in \mathbb{R}^{C \times H_{map} \times W_{map}}$ maintains the same size as the BEV features $\mathbf{\Phi}_{bev}$. The map feature is also passed through a semantic segmentation head to produce the corresponding map mask $\mathbf{M}_{map}^{sem} \in \mathbb{R}^{2 \times H_{map} \times W_{map}}$.

\subsection{Homography-Guided Pose Estimation Module}
\label{subsec: HGPR}
In this module, we introduce a high–efficiency pose regression framework. Inspired by correlation based homography estimation IHN\cite{cao2022iterative}, HCNet\cite{wang2023fine} and SSHNet\cite{yu2025sshnet}, rather than relying on attention mechanisms to implicitly infer cross–modal correspondences, we adopt an explicit strategy: using the estimated homography to guide the correlation formation between the BEV and the map.

\textbf{Feature Fusion.} We use a Siamese ResNet\cite{he2016deep} encoder to extract features for homography estimation. The input semantic features $\mathbf{\Phi}_{BEV}$ and $\mathbf{\Phi}_{Map}$ are downsampled by $4\times$ and projected to homography features via a $1\times1$ convolution. We denote these two features as $\mathbf{F}_{BEV} \in \mathbb{R}^{D \times H' \times W'}$ and $\mathbf{F}_{Map} \in \mathbb{R}^{D \times H' \times W'}$, where $H',W' = \frac{H}{4}, \frac{W}{4}$. To get fused feature, We form a dense correlation volume $\mathbf{C} \in \mathbb{R}^{H' \times W' \times H' \times W'}$ by applying dot-product matching between all spatial
locations of $\mathbf{F}_{BEV}$ and $\mathbf{F}_{Map}$:
\begin{equation}
   \mathbf{C}_{ijkl} = \mathrm{ReLU}(F_{BEV}(i,j)^{\top}F_{Map}(k,l)). 
\end{equation}
Furthermore, we apply average pooling on the correlation volume to $\mathbf{C}_{1/2} \in \mathbb{R}^{H' \times W' \times \frac{H'}{2} \times \frac{W'}{2}}$ for enlarging the receptive field. $\mathbf{C}$ and $\mathbf{C}_{1/2}$ serve as repositories of all potential correlation information required during the pose regression stage, effectively functioning as a feature warehouse.

\textbf{Iterative Pose Refinement.} Before each iteration, we crop from the precomputed correlation warehouse according to the homography estimated in the previous iteration to obtain the fused features required for the next update. Specifically, the Pixel Motion Block maintains the projected coordinates $\mathbf{X} \in \mathbb{R}^{2 \times H' \times W'}$ of $\mathbf{F}_{{BEV}}$ onto $\mathbf{F}_{{Map}}$ via the current estimated homography $\mathbf{H}_{k-1}$. These coordinates $\mathbf{X}$ are updated once before each iteration. During the cropping step, we center a fixed radius $r$ at each location in $X$ and extract a local square patch of correlations, yielding the fused feature for the $k$-th iteration $\mathbf{S}_k \in \mathbb{R}^{H' \times W' \times (2r+1) \times (2r+1)}$:
\[
\mathbf{S}_k = \{ C(u, v) \mid (u, v) \in \mathcal{N}(X_k, r) \}.
\]
Similarly, we extract the downsampled fused features $\mathbf{S}^{1/2}_k$ from $\mathbf{C}_{1/2}$ using the same procedure.

The fused features $\mathbf{S}_k$ and $\mathbf{S}^{1/2}_k$ are concatenated and fed into the pose decoder. Instead of directly predicting the 3-DoF pose $(x, y, \theta)$, the pose decoder outputs the corner displacement $\mathbf{\Delta D}_k$ for the current iteration. The estimated displacement $\mathbf{\Delta D}_k$ is then converted into the homography matrix $\mathbf{H}_k$ via Direct Linear Transformation (DLT)\cite{abdel2015direct}, which is used to update the projected coordinates $\mathbf{X}$ for the next iteration. Moreover, the estimated $\mathbf{H}_k$ further induces a geometric mapping allowing us to recover the pose parameters at this iteration.

\subsection{Mapping 3D Pose by Homography matrix}
\label{subsec: mapping}

We replace direct pose regression with a homography-based projection strategy.  This approach constrains the pose
within a feasible solution space, thereby improving estimation accuracy and convergence speed. The homography establishes the pixel-wise mapping between two planes as
\begin{equation}
    s
    \begin{bmatrix}
        u' \\ v' \\ 1
    \end{bmatrix}
    =
    \mathbf{H}
    \begin{bmatrix}
        u \\ v \\ 1
    \end{bmatrix},
    \quad
    \mathbf{H} =
    \begin{bmatrix}
        h_{11} & h_{12} & h_{13} \\
        h_{21} & h_{22} & h_{23} \\
        h_{31} & h_{32} & h_{33}
    \end{bmatrix},
\end{equation}
where $(u,v)$ and $(u',v')$ denote corresponding pixel coordinates in the BEV and map, and $s$ is a scale factor.

Given $\mathbf{H}$, the vehicle position on the map is obtained by projecting the BEV grid center $(u_c, v_c)$:
\begin{equation}
    s
    \begin{bmatrix}
        u_c' \\ v_c' \\ 1
    \end{bmatrix}
    =
    \mathbf{H}
    \begin{bmatrix}
        u_c \\ v_c \\ 1
    \end{bmatrix}.
\end{equation}
To estimate the heading angle, we additionally project an auxiliary point $(u_c, v_c+\Delta v)$ using the same transformation, yielding $(u_a', v_a')$. The line connecting $(u_c', v_c')$ and $(u_a', v_a')$ represents the forward direction on the map, from which the heading angle $\theta$ is computed as
\begin{equation}
    \theta = \arctan2(v_a' - v_c',\, u_a' - u_c') - \arctan2(\Delta v, 0).
\end{equation}

Through this geometric procedure, both the vehicle position $(x, y)$ and orientation $\theta$ can be consistently derived from the predicted homography matrix $\mathbf{H}$.

\subsection{Loss Function}
\label{subsec: LF}

To train the network in an end-to-end manner, we design a hybrid loss composed of two major components: a semantic loss and a pose regression loss. 

We leverage the semantic information of the SD map as supervision signals. Specifically, the predicted BEV semantic mask is encouraged to match the ground-truth binary map patch, while the rasterized map prediction is supervised in an identical manner. A binary cross-entropy loss is employed:
\begin{equation}
\mathcal{L}_{\text{sem}}
= \sum_{i} \mathrm{BCE}\!\left(\mathbf{M}_{i}^{sem}, \mathbf{M}_{i}^{GT}\right).
\end{equation}
where $i \in \{\text{bev}, \text{map}\}$ denotes the modality type, $\mathbf{M}_{i}^{sem}$ denotes the predicted semantic mask and $\mathbf{M}_{i}^{GT}$ is the corresponding ground-truth binary map obtained from the SD map. 

For pose regression module, we supervise the displacement and orientation using a combined regression loss. Translation is penalized by an L2 loss, while rotation is penalized by an L1 loss for robust angular regression:
\begin{equation}
\mathcal{L}_{\text{pose}} 
= \lambda_{\text{trans}}\left\| \hat{\mathbf{t}} - \mathbf{t} \right\|_2^2
+ \lambda_{\text{ori}}\left\| \hat{\theta} - \theta \right\|_1,
\end{equation}
where $(\hat{\mathbf{t}}, \hat{\theta})$ represent the predicted translation and yaw, and $(\mathbf{t}, \theta)$ are the corresponding ground-truth values.

The final training objective is a weighted sum of the above components:
\begin{equation}
\mathcal{L} 
= \lambda_{\text{sem}} \mathcal{L}_{\text{semantic}}
+ \mathcal{L}_{\text{pose}}.
\end{equation}
$\lambda_{\text{sem}}$, $\lambda_{\text{trans}}$ and $\lambda_{\text{ori}}$ are factors to balance semantic supervision and pose refinement.

\section{Experiments}

\begin{table*}[ht]
  \caption{Localization results on nuScenes dataset. * denotes the data is taken directly from the original paper. $^1$ $C$: Camera, $L$: Lidar.}
  \centering
  \setlength{\tabcolsep}{1.2mm}{
  \begin{tabular}{lcccccccccccc}
    \toprule
    \multicolumn{1}{c}{\multirow{2.5}*{Method}} &
    \multirow{2.5}*{Input$^1$} &
    \multicolumn{4}{c}{\centering Position Recall@$Xm$ $\uparrow$} &
    \multicolumn{4}{c}{\centering Orientation Recall@$X^\circ$ $\uparrow$} &
    \multirow{2.5}*{APE$(m)$ $\downarrow$} & 
    \multirow{2.5}*{AOE$(^\circ)$ $\downarrow$} \\
    \cmidrule(r){3-6} \cmidrule(r){7-10}
    & &
    \multicolumn{1}{c}{\centering $1m$} &
    \multicolumn{1}{c}{\centering $2m$} &
    \multicolumn{1}{c}{\centering $5m$} &
    \multicolumn{1}{c}{\centering $10m$} &
    \multicolumn{1}{c}{\centering $1^\circ$} & 
    \multicolumn{1}{c}{\centering $2^\circ$} &
    \multicolumn{1}{c}{\centering $5^\circ$} &
    \multicolumn{1}{c}{\centering $10^\circ$} \\
    \midrule
    SegLocNet-road*\cite{zhou2025seglocnet} & $C+L$ & $35.63$ & $57.98$ & $74.55$ & $80.09$ & $38.58$ & $64.98$ & $83.62$ & $87.40$ & $8.15$ & $19.68$ \\
    SegLocNet-drivable*\cite{zhou2025seglocnet} & $C+L$ & $59.08$ & $76.04$ & $84.25$ & $86.86$ & $63.19$ & $84.55$ & $91.02$ & $93.00$ & ${5.30}$ & ${10.11}$ \\
    
    \midrule
    OrienterNet\cite{sarlin2023orienternet} & $C$ & $15.74$ & $33.27$ & $49.86$ & $60.10$ & $31.69$ & $45.81$ & $61.23$ & $72.44$ & $15.47$ & $26.43$ \\
    U-BEV*\cite{camiletto2024u}& $C$ & $16.89$ & $41.60$ & $71.33$ & $83.46$ & $-$ & $-$ & $-$ & $-$ & $-$ & $-$ \\
    MapLocNet One-Stage*\cite{wu2024maplocnet} & $C$ & $16.32$ & $40.56$ & $74.27$ & $89.69$ & $53.71$ & $78.13$ & $94.49$ & $98.05$ & $-$ & $-$ \\    
    MapLocNet*\cite{wu2024maplocnet} & $C$ & $20.10$ & $45.54$ & $77.70$ & $91.89$ & $58.61$ & $84.10$ & $96.23$ & $98.62$ & $-$ & $-$ \\
    HOLO-CA(ours) & $C$ & $21.47$ & $46.70$ & $77.71$ & $90.02$ & $39.52$ & $76.13$ & $94.16$ & $98.41$ & $4.31$ & $1.78$ \\
    HOLO-road(ours) & $C$ & $30.40$ & $54.36$ & $82.11$ & $92.46$ & $69.76$ &
    $88.33$ & $\textbf{98.14}$ & $99.18$ & $3.76$ & $1.06$\\
    HOLO-drivable(ours) & $C$ & $\textbf{36.41}$ & $\textbf{61.21}$ & $\textbf{84.10}$ & $\textbf{93.09}$ & $\textbf{73.74}$ & $\textbf{91.51}$ & $97.35$ & $\textbf{99.20}$ & $\textbf{3.37}$ & $\textbf{1.02}$\\
    \bottomrule
  \end{tabular}}
  \label{tab:localization_result}
\end{table*}

\begin{figure*}[ht]
  \centering
  \includegraphics[width=0.9\linewidth]{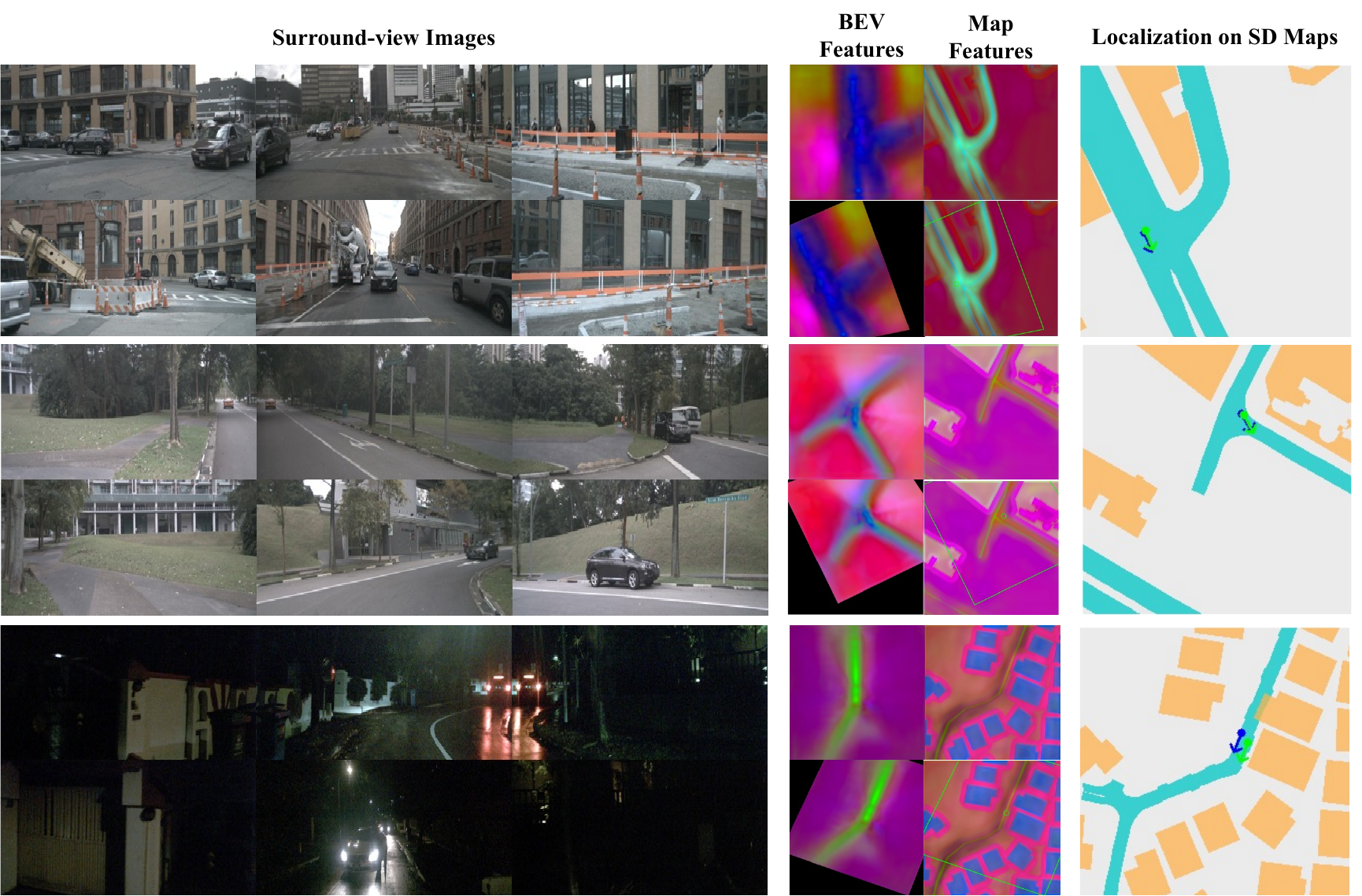}
  \caption{The localization results on nuScenes dataset. The second row illustrates localization results during a left-turn maneuver, and the third row shows the localization performance during nighttime driving. The third column shows the BEV features before and after warping, and the fourth column displays the SD map features with the warped BEV positions. Shown in the fifth row is the SD map with drivable areas and buildings. The green arrows indicate the ground-truth vehicle poses, while the blue arrows represent the estimated poses.}
  \label{fig:result_viz}
\end{figure*}

\subsection{Datasets}
We conduct training and evaluation on the nuScenes\cite{caesar2020nuscenes} autonomous driving dataset. NuScenes contains 1000 real-world driving scenes collected across Boston and Singapore. Following the official data split, we adopt 700 scenes for training and use the remaining 150 scenes as the validation set to benchmark our model’s performance. 

Since nuScenes does not provide SD maps, we retrieve the corresponding regions from OSM and align them to the nuScenes coordinate system. Unlike prior work \cite{Blos-BEV, wu2024maplocnet, zhou2025seglocnet} that aligns OSM maps in EPSG:3857 and requires manual scaling for the Boston area, we directly transform WGS84 coordinates into the ENU frame used by nuScenes. Specifically, we choose a fixed reference origin per region, convert WGS84 to ECEF, and then map ECEF to a local ENU frame. A small constant drift correction is applied to ensure precise alignment, particularly for the Singapore scenes where minor GPS drift exists in the raw data. As a results, well-aligned SD maps were obtained for each keyframe across 1,000 scenes, thereby extending the nuScenes dataset.

\begin{table*}[ht]
  \caption{Ablation of homography estimation. Homo. denotes the model equipped with the homography estimation head. }
  \centering
  \setlength{\tabcolsep}{1.5mm}{
  \begin{tabular}{lcccccccccccc}
    \toprule
    \multicolumn{1}{c}{\multirow{2.5}*{Method}} &
    \multirow{2.5}*{Head} &
    \multicolumn{4}{c}{\centering Position Recall@$Xm$ $\uparrow$} &
    \multicolumn{4}{c}{\centering Orientation Recall@$X^\circ$ $\uparrow$} &
    \multirow{2.5}*{APE$(m)$ $\downarrow$} & 
    \multirow{2.5}*{AOE$(^\circ)$ $\downarrow$} \\
    \cmidrule(r){3-6} \cmidrule(r){7-10}
    & &
    \multicolumn{1}{c}{\centering $1m$} &
    \multicolumn{1}{c}{\centering $2m$} &
    \multicolumn{1}{c}{\centering $5m$} &
    \multicolumn{1}{c}{\centering $10m$} &
    \multicolumn{1}{c}{\centering $1^\circ$} & 
    \multicolumn{1}{c}{\centering $2^\circ$} &
    \multicolumn{1}{c}{\centering $5^\circ$} &
    \multicolumn{1}{c}{\centering $10^\circ$} \\
    \midrule
    HOLO-CA & Pose & $10.06$ & $31.18$ & $71.09$ & $88.99$ & $18.04$ & $61.01$ & $93.37$ & $\textbf{98.67}$ & $5.02$ & $2.24$ \\
    HOLO-CA & Homo. & $\textbf{21.47}$ & $\textbf{46.70}$ & $\textbf{77.71}$ & $\textbf{90.02}$ & $\textbf{39.52}$ & $\textbf{76.13}$ & $\textbf{94.16}$ & $98.41$ & $\textbf{4.31}$ & $\textbf{1.78}$ \\
    \midrule
    HOLO(one-iter) & Pose & $9.88$ & $31.24$ & $71.10$ & $89.29$ & $10.08$ & $61.27$ & $94.43$ & $98.94$ & $5.11$ & $2.19$ \\
    HOLO(one-iter) & Homo. & $\textbf{26.57}$ & $\textbf{52.07}$ & $\textbf{79.99}$ & $\textbf{90.80}$ & $\textbf{45.62}$ & $\textbf{83.02}$ & $\textbf{96.02}$ & $\textbf{99.20}$ & $\textbf{4.07}$ & $\textbf{1.49}$ \\
    \bottomrule
  \end{tabular}}
  \label{tab:ablation_head}
\end{table*}

\subsection{Implementation Details}
\textbf{Network settings.} Each of the six surround-view images is resized to \(128 \times 352\) before feature extraction. The BEV covers a \(64\,\text{m} \times 64\,\text{m}\) area centered on the vehicle, spanning \([-32\,\text{m}, 32\,\text{m}]\) along both longitudinal and lateral directions, with a spatial resolution of \(0.25\,\text{m/pixel}\). Depth values are discretized from \(4\,\text{m}\) to \(27\,\text{m}\) at \(1\,\text{m}\) intervals. The map input is a cropped \(128\,\text{m} \times 128\,\text{m}\) rasterized SD map at \(0.5\,\text{m/pixel}\) resolution. We simulate GPS noise by randomly perturbing the map patch with rotations of up to $\pm30^\circ$ and translations of up to $\pm30\,\text{m}$.

\textbf{Training details.} We implement our network in PyTorch. Training is performed on an NVIDIA RTX A6000 GPU using the AdamW\cite{loshchilov2017decoupled} optimizer with a maximum learning rate of $3.5 \times 10^{-4}$ and a weight decay of $5 \times 10^{-4}$. The batch size is set to 16, and the model is trained for 180,000 iterations. A OneCycle\cite{smith2017cyclical} learning rate scheduler is employed to adjust the learning rate throughout training. The loss weights are set as:
$\lambda_{\text{sem}} = 1000,
\lambda_{\text{trans}} = 1,
\lambda_{\text{ori}} = 10$
and the iterative pose refinement is performed for $N = 6$ iterations.

\textbf{Metrics.} We follow the evaluation metrics defined in \cite{zhou2025seglocnet,wu2024maplocnet}, 
including Recall@X\,m, Recall@X\textdegree, Absolute Position Error (APE), and Absolute Orientation Error (AOE), 
which jointly assess translational and rotational localization accuracy.
\begin{figure}[t]
  \centering
   \includegraphics[width=0.97\linewidth]{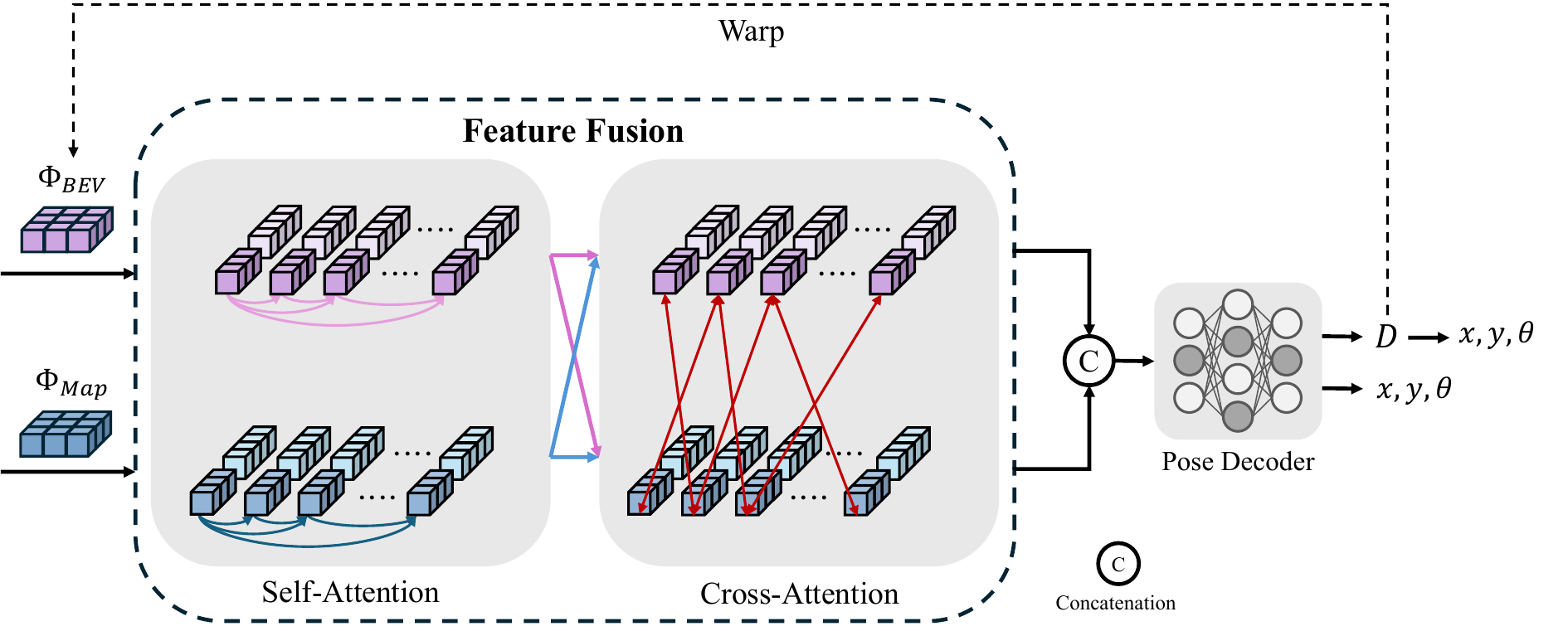}
   \caption{The overall architecture of \textbf{HOLO-CA}. The BEV feature and the map feature are fed into a cross-attention-based feature fusion module. The fused features are then concatenated and passed to the pose decoder, which can be either a homography decoder or a 3-DOF pose decoder. The iterative strategy can also be employed by direcly warpping BEV features.}
   \label{Fig:HOLO_CA}
\end{figure}
\subsection{Comparisons with Existing Methods}

We compare our proposed network HOLO with several existing methods: OrienterNet \cite{sarlin2023orienternet}, U-BEV \cite{camiletto2024u}, MapLocNet \cite{wu2024maplocnet}, SegLocNet \cite{zhou2025seglocnet}. In addition, to further validate the effectiveness of our homography-guided strategy, we implement a pose estimation network based on cross-attention feature fusion\cite{sun2021loftr,qin2022geometric}, namely HOLO-CA, as shown in \cref{Fig:HOLO_CA}. This network is also included in our comparison.

As shown in \cref{tab:localization_result}, our method outperforms existing visual localization methods in all metrics. Moreover, unlike prior works such as MapLocNet and U-BEV that rely on the \emph{drivable area} layer from nuScenes HD maps, our method achieves comparably high accuracy even when supervised only by the \emph{road} layer from SD maps, with negligible performance degradation. Specifically, our approach consistently yields substantial gains over MapLocNet in Recall@1m/2m and Recall@1$^\circ$/2$^\circ$, regardless of the supervision map format. When supervised by the \textit{road} layer, we observe improvements of \textbf{10.30\%/8.82\%} (Recall@1m/2m) and \textbf{11.15\%/4.23\%} (Recall@1$^\circ$/2$^\circ$). When using \textit{drivable area} supervision, the gains increase to \textbf{16.31\%/15.67\%} (Recall@1m/2m) and \textbf{15.13\%/7.41\%} (Recall@1$^\circ$/2$^\circ$). It is also worth noting that our HOLO-CA achieves comparable performance to MapLocNet.
\begin{table}[ht]
  \centering
  \caption{Comparison of different numbers of iterations.}
  \scalebox{0.8}{
  \setlength{\tabcolsep}{3pt}{
  \begin{tabular}{ccccccccc}
    \toprule
    \multirow{2.5}*{Iteration} &
    \multicolumn{4}{c}{\centering Position Recall@$Xm$ $\uparrow$} &
    \multicolumn{4}{c}{\centering Orientation Recall@$X^\circ$ $\uparrow$} \\
    \cmidrule(r){2-5} \cmidrule(r){6-9}
    & 
    \multicolumn{1}{c}{\centering $1m$} &
    \multicolumn{1}{c}{\centering $2m$} &
    \multicolumn{1}{c}{\centering $5m$} &
    \multicolumn{1}{c}{\centering $10m$} &
    \multicolumn{1}{c}{\centering $1^\circ$} & 
    \multicolumn{1}{c}{\centering $2^\circ$} &
    \multicolumn{1}{c}{\centering $5^\circ$} &
    \multicolumn{1}{c}{\centering $10^\circ$} \\
    \midrule
    1 & $26.57$ & $52.07$ & $79.99$ & $90.80$ & $45.62$ & $83.02$ & $96.02$ & $99.20$ \\
    2 & $31.64$ & $56.21$ & $82.54$ & $91.78$ & $67.37$ & $87.80$ & $\textbf{98.14}$ & $98.94$ \\
    6 & $\textbf{36.41}$ & $\textbf{61.21}$ & $\textbf{84.10}$ & $\textbf{93.09}$ & $\textbf{73.74}$ & $\textbf{91.51}$ & $97.35$ & $\textbf{99.20}$ \\
    \bottomrule
  \end{tabular}}}
  \label{tab:exp_iter}
\end{table}

\begin{table}[t]
  \caption{Comparison of the average performance variation brought by each iteration across different methods. * denotes the results calculated by the data directly taken from original paper.}
  \label{tab:exp_performance_comp}
  \centering
  \begin{tabular}{@{}lcc@{}}
    \toprule
    Method & $\Delta \text{GFLOPs} \downarrow$ & $ \Delta \text{FPS} \uparrow$\\
    \midrule
    MapLocNet*\cite{wu2024maplocnet} & $+3.01$ & $-25.41\%$\\
    HOLO-CA(ours) & $+9.85$ & $-26.47\%$\\
    HOLO(ours) & $\textbf{+0.98}$ & $\textbf{-5.62}\%$\\
    \bottomrule
  \end{tabular}
\end{table}

We further compare HOLO to a multimodal method SegLocNet. While our method is slightly lower in Recall@1m/2m, it surpasses multimodal approaches across all other metrics. This advantage likely stems from LiDAR inputs which narrow the domain gap and boost perception quality. 

\subsection{Ablation Study}

\begin{figure}[t]
    \centering
    \begin{subfigure}[b]{0.45\linewidth}
        \centering
        \includegraphics[width=\textwidth]{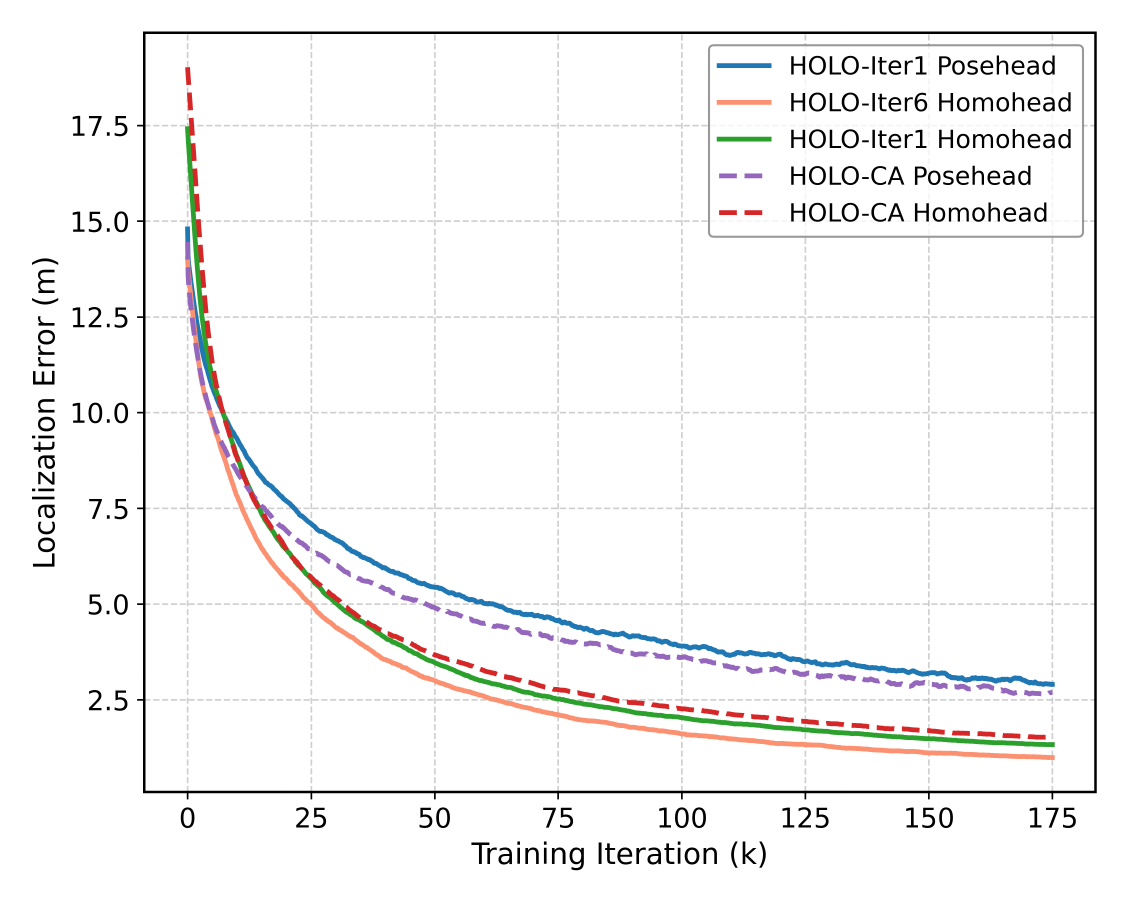}
        \caption{Localization Error in Training}
        \label{fig:sub1}
    \end{subfigure}
    \hfill
    \begin{subfigure}[b]{0.45\linewidth}
        \centering
        \includegraphics[width=\textwidth]{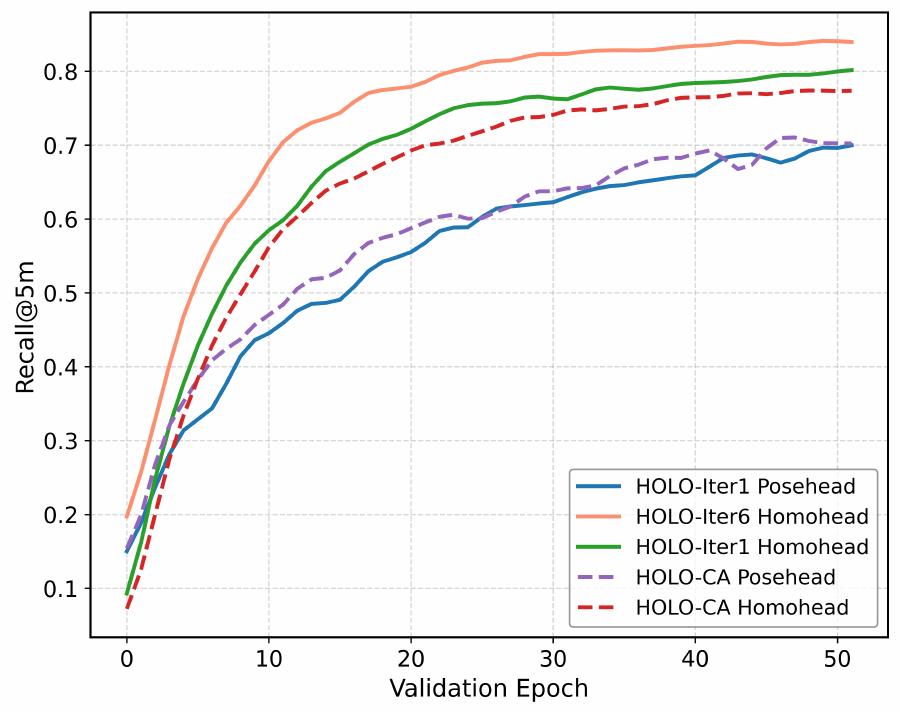}
        \caption{$Recall@5m$ in Validation}
        \label{fig:sub2}
    \end{subfigure}
    \begin{subfigure}[b]{0.45\linewidth}
        \centering
        \includegraphics[width=\textwidth]{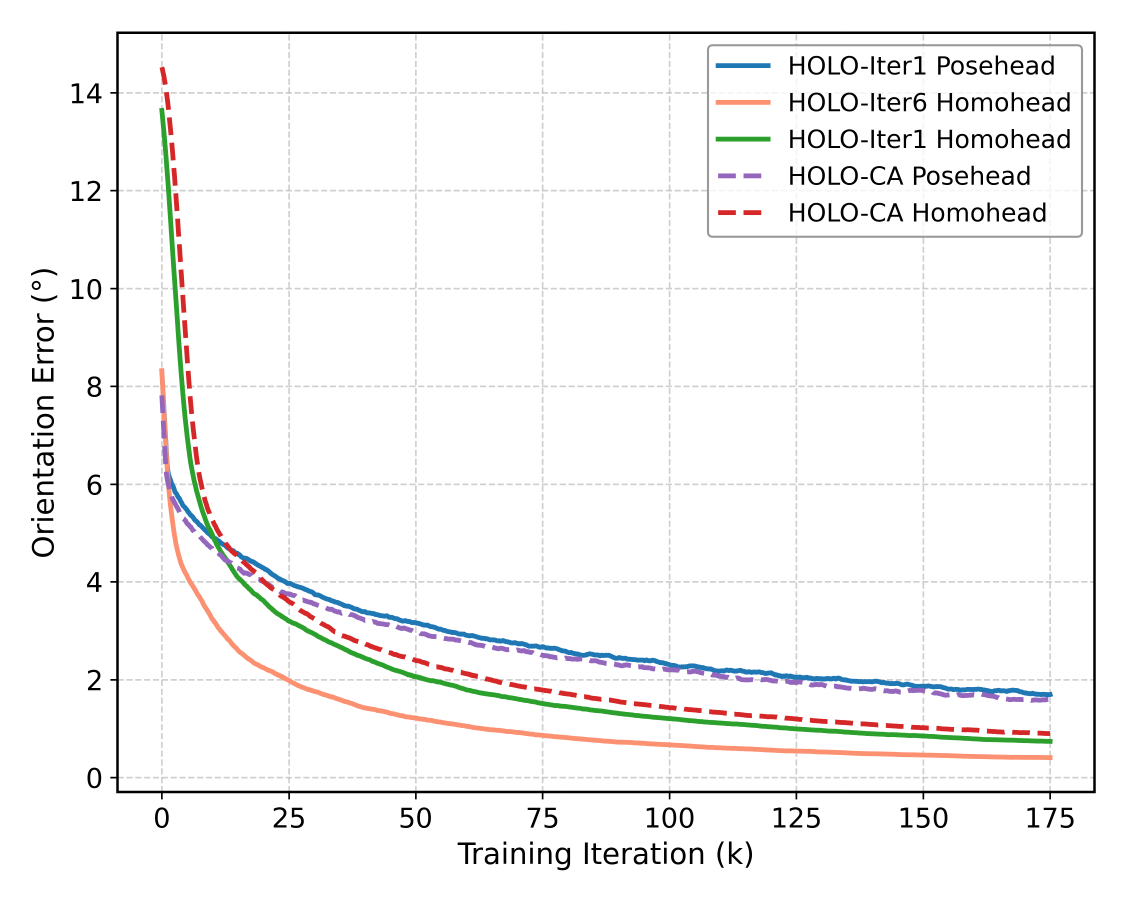}
        \caption{Orientation Error in Training}
        \label{fig:sub3}
    \end{subfigure}
    \hfill
    \begin{subfigure}[b]{0.45\linewidth}
        \centering
        \includegraphics[width=\textwidth]{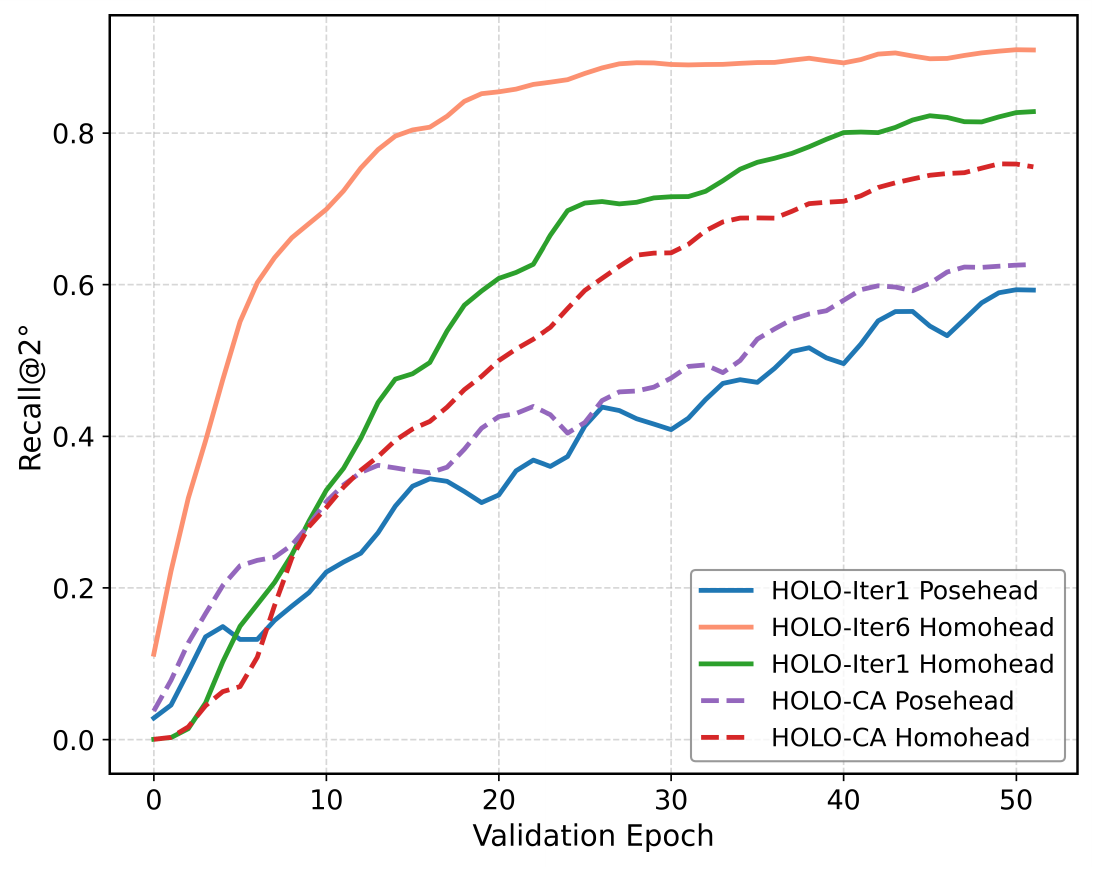}
        \caption{$Recall@2^\circ$ in Validation}
        \label{fig:sub4}
    \end{subfigure}
    \caption{Training and validation curves of different methods.}
    \label{fig:all}
\end{figure}

\textbf{Ablation of Homography estimation.} 
To verify the effectiveness of constraining the pose within a feasible region through homography, we replace the final homography regression head in both HOLO and HOLO-CA with a 3-DOF pose regression head, while disabling the iterative optimization strategy.
As shown in \cref{tab:ablation_head}, regardless of the feature fusion strategy used, substituting the pose regression head with the homography regression head leads to significant improvements across all metrics.
Furthermore, as illustrated in \cref{fig:all}, introducing homography to restrict the pose regression space enables faster convergence and yields lower errors compared to directly regressing poses, which also indicates higher training efficiency.

\textbf{Ablation of Iterative Strategy.} 
To further improve the pose estimation accuracy, we adopt an iterative refinement strategy. \cref{tab:exp_iter} reports the localization performance of our method under different numbers of iterations. We observe that increasing the number of iterations significantly boosts metrics such as Recall@1m/2m and Recall@1°/2°. This demonstrates that iterative refinement enables our model to produce more precise pose estimates.

\textbf{Analysis of Feature Fusion Approaches.} 
From \cref{tab:ablation_head}, we observe that our homography-guided fusion strategy achieves better performance than the cross-attention-based fusion approach. Furthermore, as we adopt an iterative process to refine pose estimation, \cref{tab:exp_performance_comp} reports the average changes in GFLOPs and FPS per iteration for different methods. We find that our method only updates the sampling positions of pre-computed correlations rather than recomputing attention-based feature fusion. As a result, each additional iteration introduces minimal computational overhead—only one extra pass through the pose decoder—while methods based on attention incur significantly higher costs. This demonstrates that our fusion strategy is both training-efficient and inference-efficient.

\begin{table}[htbp]
  \centering
  \caption{Experimental results of robust analysis.}
  \setlength{\tabcolsep}{2.5pt}
  \begin{subtable}[t]{0.95\linewidth}
    \centering
    \caption{Localization results under 40\,m noise level.}
    \scalebox{0.8}{
    \begin{tabular}{cccccc}
      \toprule
      \multirow{2.5}*{Noise (m)} &
      \multicolumn{4}{c}{Position Recall@$Xm$ $\uparrow$} &
      \multirow{2.5}*{APE (m) $\downarrow$} \\
      \cmidrule(r){2-5}
      & $1$m & $2$m & $5$m & $10$m \\
      \midrule
      40 & 18.24 & 44.84 & 76.87 & 89.89 & 5.01 \\
      \bottomrule
    \end{tabular}}
    \label{tab:exp_noise_level}
  \end{subtable}
  \vspace{1em} 
  \begin{subtable}[t]{0.95\linewidth}
    \centering
    \caption{Results under different BEV resolutions.}
    \scalebox{0.8}{
    \begin{tabular}{ccccccccc}
      \toprule
      \multirow{2.5}*{Grid(\textit{mpp})} &
      \multicolumn{4}{c}{Position Recall@$Xm$ $\uparrow$} &
      \multicolumn{4}{c}{Orientation Recall@$X^\circ$ $\uparrow$} \\
      \cmidrule(r){2-5} \cmidrule(r){6-9}
      &
      $1$m & $2$m & $5$m & $10$m &
      $1^\circ$ & $2^\circ$ & $5^\circ$ & $10^\circ$ \\
      \midrule
      \textbf{0.25} & 36.41 & 61.21 & 84.10 & 93.09 & 73.74 & 91.51 & 97.35 & 99.20 \\
      0.5 & 36.10 & 61.52 & 83.22 & 92.37 & 71.09 & 90.72 & 98.14 & 99.20 \\
      \bottomrule
    \end{tabular}}
    \label{tab:exp_cross_resolution}
  \end{subtable}
  \label{tab:exp_robust}
\end{table}

\subsection{Extended Evaluation and Robustness Analysis}

\textbf{Robust Analysis.} To evaluate the robustness of our model, we design two sets of experiments. First, since our method is built upon homography-based modeling, we fix the map resolution to 0.5 \textit{mpp} and test the performance under different BEV input resolutions. As shown in \cref{tab:exp_cross_resolution}, the results remain largely stable across varying BEV resolutions, indicating that our model can effectively handle cross-resolution inputs.

In addition, we increase the simulated noise range from 30m to 40m, creating a more challenging scenario where the input map patch may not fully overlap with the perceived BEV region. As shown in \cref{tab:exp_noise_level}, although performance slightly degrades under this setting, our method still achieves comparable results to MapLocNet under the 30m noise condition, demonstrating strong robustness to environmental perturbations.

\begin{table}[ht]
  \caption{IoU performance with/without Homography Learning.}
  \label{tab:exp_seg_IoU}
  \centering
  \begin{tabular}{@{}lcc@{}}
    \toprule
    Method & $\text{Drivable Area} \uparrow$ & $\text{Buildings} \uparrow$\\
    \midrule
    \textit{w/o} Homo. & $76.52$ & $68.41$\\
    \textit{w/} Homo. & $79.89$ & $69.43$\\
    \bottomrule
  \end{tabular}
\end{table}

\textbf{Segmentation Evaluation.} We further evaluate the semantic segmentation performance of the BEV perception module when jointly trained with homography learning, as shown in \cref{tab:exp_seg_IoU}. For comparison, we also train the BEV perception network independently using the same grid configuration as in the joint training setup. The results show that, under joint training, the segmentation IoU of the driveable area improves by 3.37\%, and that of building improves by 1.02\% compared to the standalone model. We infer that the semantic cues from the map input facilitate a more consistent alignment of BEV features through the homography-based learning, which in turn enhances the segmentation capability of the BEV perception module.

\textbf{Inference Time.} We run our model on an Intel(R) Xeon(R) Platinum 8336C CPU @ 2.30GHz and an Nvidia RTX A6000 GPU. The two-iteration version achieves \textbf{25.6} FPS, while the six-iteration version achieves \textbf{19.9} FPS.

\section{Conclusion}
We have proposed novel homography-guided pose estimator network for fine-grained visual localization between multi-view images and
standard-definition maps, named HOLO. HOLO leverages semantic cues to bridge BEV and map representations, integrating homography constraints into the pose estimation network to overcome the limitations of current regression methods. By guiding feature fusion with homography, our method improves both training and inference efficiency. Replacing the pose regression head with a homography estimation head significantly enhances localization accuracy and convergence speed. We also extend the nuScenes dataset with SD map data, which will be released publicly to promote research on localization with SD maps. In future work, we plan to incorporate temporal perception information to further improve localization performance.

\section*{Acknowledgement}
This work was supported by the National Natural Science Foundation of China (No.62133002) and the State Key Laboratory of Autonomous Intelligent Unmanned Systems, Beijing Institute of Technology.

{
    \small
    \bibliographystyle{ieeenat_fullname}
    \bibliography{main}

@String(CVPR= {IEEE Conf. Comput. Vis. Pattern Recog.})

@String(ICIP = {IEEE Int. Conf. Image Process.})

@String(CVPR  = {CVPR})

@String(ICIP  = {ICIP})

@inproceedings{samano2020you,
  title={You are here: Geolocation by embedding maps and images},
  author={Samano, Noe and Zhou, Mengjie and Calway, Andrew},
  booktitle={European Conference on Computer Vision},
  pages={502--518},
  year={2020},
  organization={Springer}
}

@inproceedings{zhou2021efficient,
  title={Efficient localisation using images and OpenStreetMaps},
  author={Zhou, Mengjie and Chen, Xieyuanli and Samano, Noe and Stachniss, Cyrill and Calway, Andrew},
  booktitle={2021 IEEE/RSJ International Conference on Intelligent Robots and Systems (IROS)},
  pages={5507--5513},
  year={2021},
  organization={IEEE}
}

@inproceedings{sarlin2023orienternet,
  title={Orienternet: Visual localization in 2d public maps with neural matching},
  author={Sarlin, Paul-Edouard and DeTone, Daniel and Yang, Tsun-Yi and Avetisyan, Armen and Straub, Julian and Malisiewicz, Tomasz and Bulo, Samuel Rota and Newcombe, Richard and Kontschieder, Peter and Balntas, Vasileios},
  booktitle={Proceedings of the IEEE/CVF Conference on Computer Vision and Pattern Recognition},
  pages={21632--21642},
  year={2023}
}

@article{sarlin2023snap,
  title={Snap: Self-supervised neural maps for visual positioning and semantic understanding},
  author={Sarlin, Paul-Edouard and Trulls, Eduard and Pollefeys, Marc and Hosang, Jan and Lynen, Simon},
  journal={Advances in Neural Information Processing Systems},
  volume={36},
  pages={7697--7729},
  year={2023}
}

@inproceedings{wu2024maplocnet,
  title={Maplocnet: Coarse-to-fine feature registration for visual re-localization in navigation maps},
  author={Wu, Hang and Zhang, Zhenghao and Lin, Siyuan and Mu, Xiangru and Zhao, Qiang and Yang, Ming and Qin, Tong},
  booktitle={2024 IEEE/RSJ International Conference on Intelligent Robots and Systems (IROS)},
  pages={13198--13205},
  year={2024},
  organization={IEEE}
}

@article{zhou2025seglocnet,
  title={SegLocNet: Multimodal Localization Network for Autonomous Driving via Bird's-Eye-View Segmentation},
  author={Zhou, Zijie and Qi, Zhangshuo and Cheng, Luqi and Xiong, Guangming},
  journal={arXiv preprint arXiv:2502.20077},
  year={2025}
}

@inproceedings{zhao2021deep,
  title={Deep lucas-kanade homography for multimodal image alignment},
  author={Zhao, Yiming and Huang, Xinming and Zhang, Ziming},
  booktitle={Proceedings of the IEEE/CVF conference on computer vision and pattern recognition},
  pages={15950--15959},
  year={2021}
}

@inproceedings{cao2022iterative,
  title={Iterative deep homography estimation},
  author={Cao, Si-Yuan and Hu, Jianxin and Sheng, Zehua and Shen, Hui-Liang},
  booktitle={Proceedings of the IEEE/CVF conference on computer vision and pattern recognition},
  pages={1879--1888},
  year={2022}
}

@inproceedings{cao2023recurrent,
  title={Recurrent homography estimation using homography-guided image warping and focus transformer},
  author={Cao, Si-Yuan and Zhang, Runmin and Luo, Lun and Yu, Beinan and Sheng, Zehua and Li, Junwei and Shen, Hui-Liang},
  booktitle={Proceedings of the IEEE/CVF Conference on Computer Vision and Pattern Recognition},
  pages={9833--9842},
  year={2023}
}

@inproceedings{zhang2024scpnet,
  title={SCPNet: Unsupervised Cross-modal Homography Estimation via Intra-modal Self-supervised Learning},
  author={Zhang, Runmin and Ma, Jun and Cao, Si-Yuan and Luo, Lun and Yu, Beinan and Chen, Shu-Jie and Li, Junwei and Shen, Hui-Liang},
  booktitle={European Conference on Computer Vision},
  pages={460--477},
  year={2024},
  organization={Springer}
}

@article{song2024unsupervised,
  title={Unsupervised homography estimation on multimodal image pair via alternating optimization},
  author={Song, Sanghyeob and Lew, Jaihyun and Jang, Hyemi and Yoon, Sungroh},
  journal={Advances in Neural Information Processing Systems},
  volume={37},
  pages={61306--61327},
  year={2024}
}

@inproceedings{yu2025sshnet,
  title={SSHNet: Unsupervised Cross-modal Homography Estimation via Problem Reformulation and Split Optimization},
  author={Yu, Junchen and Cao, Si-Yuan and Zhang, Runmin and Zhang, Chenghao and Yu, Zhu and Chen, Shujie and Yang, Bailin and Shen, Hui-Liang},
  booktitle={Proceedings of the Computer Vision and Pattern Recognition Conference},
  pages={16685--16694},
  year={2025}
}

@inproceedings{camiletto2024u,
  title={U-bev: Height-aware bird’s-eye-view segmentation and neural map-based relocalization},
  author={Camiletto, Andrea Boscolo and Bochicchio, Alfredo and Liniger, Alexander and Dai, Dengxin and Gawel, Abel},
  booktitle={2024 IEEE/RSJ International Conference on Intelligent Robots and Systems (IROS)},
  pages={5597--5604},
  year={2024},
  organization={IEEE}
}

@inproceedings{chen2024map,
  title={Map-relative pose regression for visual re-localization},
  author={Chen, Shuai and Cavallari, Tommaso and Prisacariu, Victor Adrian and Brachmann, Eric},
  booktitle={Proceedings of the IEEE/CVF Conference on Computer Vision and Pattern Recognition},
  pages={20665--20674},
  year={2024}
}

@article{wang2023fine,
  title={Fine-grained cross-view geo-localization using a correlation-aware homography estimator},
  author={Wang, Xiaolong and Xu, Runsen and Cui, Zhuofan and Wan, Zeyu and Zhang, Yu},
  journal={Advances in Neural Information Processing Systems},
  volume={36},
  pages={5301--5319},
  year={2023}
}

@article{zhang2025bev,
  title={Bev-locator: An end-to-end visual semantic localization network using multi-view images},
  author={Zhang, Zhihuang and Xu, Meng and Zhou, Wenqiang and Peng, Tao and Li, Liang and Poslad, Stefan},
  journal={Science China Information Sciences},
  volume={68},
  number={2},
  pages={122106},
  year={2025},
  publisher={Springer}
}

@inproceedings{he2024egovm,
  title={Egovm: Achieving precise ego-localization using lightweight vectorized maps},
  author={He, Yuzhe and Liang, Shuang and Rui, Xiaofei and Cai, Chengying and Wan, Guowei},
  booktitle={2024 IEEE/RSJ International Conference on Intelligent Robots and Systems (IROS)},
  pages={12248--12255},
  year={2024},
  organization={IEEE}
}

@InProceedings{Zhao2024towards,
  author={Zhao, Lili and Liu, Zhili and Yin, Qian and Yang, Lei and Guo, Meng},
  booktitle={2024 IEEE International Conference on Image Processing (ICIP)}, 
  title={Towards Robust Visual Localization Using Multi-View Images and HD Vector Map}, 
  year={2024},
  pages={814-820},
}

@ARTICLE{Haklay2008open,
  author={Haklay, Mordechai and Weber, Patrick},
  journal={IEEE Pervasive Computing}, 
  title={OpenStreetMap: User-Generated Street Maps}, 
  year={2008},
  volume={7},
  number={4},
  pages={12-18}
}

@inproceedings{philion2020lift,
  title={Lift, splat, shoot: Encoding images from arbitrary camera rigs by implicitly unprojecting to 3d},
  author={Philion, Jonah and Fidler, Sanja},
  booktitle={European conference on computer vision},
  pages={194--210},
  year={2020},
  organization={Springer}
}

@article{vgg,
  title={Very deep convolutional networks for large-scale image recognition},
  author={Simonyan, Karen and Zisserman, Andrew},
  journal={arXiv preprint arXiv:1409.1556},
  year={2014}
}

@inproceedings{lentsch2023slicematch,
  title={Slicematch: Geometry-guided aggregation for cross-view pose estimation},
  author={Lentsch, Ted and Xia, Zimin and Caesar, Holger and Kooij, Julian FP},
  booktitle={Proceedings of the IEEE/CVF Conference on Computer Vision and Pattern Recognition},
  pages={17225--17234},
  year={2023}
}

@article{xia2023convolutional,
  title={Convolutional cross-view pose estimation},
  author={Xia, Zimin and Booij, Olaf and Kooij, Julian FP},
  journal={IEEE Transactions on Pattern Analysis and Machine Intelligence},
  volume={46},
  number={5},
  pages={3813--3831},
  year={2023},
  publisher={IEEE}
}

@inproceedings{yuan2024cross,
  title={Cross-attention between satellite and ground views for enhanced fine-grained robot geo-localization},
  author={Yuan, Dong and Maire, Frederic and Dayoub, Feras},
  booktitle={Proceedings of the IEEE/CVF Winter Conference on Applications of Computer Vision},
  pages={1249--1256},
  year={2024}
}

@InProceedings{Dong_2025_CVPR,
    author    = {Dong, Siyan and Wang, Shuzhe and Liu, Shaohui and Cai, Lulu and Fan, Qingnan and Kannala, Juho and Yang, Yanchao},
    title     = {Reloc3r: Large-Scale Training of Relative Camera Pose Regression for Generalizable, Fast, and Accurate Visual Localization},
    booktitle = {Proceedings of the IEEE/CVF Conference on Computer Vision and Pattern Recognition (CVPR)},
    month     = {June},
    year      = {2025},
    pages     = {16739-16752}
}

@INPROCEEDINGS{Blos-BEV,
  author={Wu, Hang and Zhang, Zhenghao and Lin, Siyuan and Qin, Tong and Pan, Jin and Zhao, Qiang and Xu, Chunjing and Yang, Ming},
  booktitle={2024 IEEE Intelligent Vehicles Symposium (IV)}, 
  title={BLOS-BEV: Navigation Map Enhanced Lane Segmentation Network, Beyond Line of Sight}, 
  year={2024},
  pages={3212-3219},
}

@inproceedings{caesar2020nuscenes,
  title={nuscenes: A multimodal dataset for autonomous driving},
  author={Caesar, Holger and Bankiti, Varun and Lang, Alex H and Vora, Sourabh and Liong, Venice Erin and Xu, Qiang and Krishnan, Anush and Pan, Yu and Baldan, Giancarlo and Beijbom, Oscar},
  booktitle={Proceedings of the IEEE/CVF conference on computer vision and pattern recognition},
  pages={11621--11631},
  year={2020}
}

@inproceedings{tan2019efficientnet,
  title={Efficientnet: Rethinking model scaling for convolutional neural networks},
  author={Tan, Mingxing and Le, Quoc},
  booktitle={International conference on machine learning},
  pages={6105--6114},
  year={2019},
  organization={PMLR}
}

@article{loshchilov2017decoupled,
  title={Decoupled weight decay regularization},
  author={Loshchilov, Ilya and Hutter, Frank},
  journal={arXiv preprint arXiv:1711.05101},
  year={2017}
}

@inproceedings{smith2017cyclical,
  title={Cyclical learning rates for training neural networks},
  author={Smith, Leslie N},
  booktitle={2017 IEEE winter conference on applications of computer vision (WACV)},
  pages={464--472},
  year={2017},
  organization={IEEE}
}

@article{abdel2015direct,
  title={Direct linear transformation from comparator coordinates into object space coordinates in close-range photogrammetry},
  author={Abdel-Aziz, Yousset I and Karara, Hauck Michael and Hauck, Michael},
  journal={Photogrammetric engineering \& remote sensing},
  volume={81},
  number={2},
  pages={103--107},
  year={2015},
  publisher={Elsevier}
}

@article{li2024bevformer,
  title={Bevformer: learning bird’s-eye-view representation from lidar-camera via spatiotemporal transformers},
  author={Li, Zhiqi and Wang, Wenhai and Li, Hongyang and Xie, Enze and Sima, Chonghao and Lu, Tong and Yu, Qiao and Dai, Jifeng},
  journal={IEEE Transactions on Pattern Analysis and Machine Intelligence},
  volume={47},
  number={3},
  pages={2020--2036},
  year={2024},
  publisher={IEEE}
}

@inproceedings{li2023fb,
  title={Fb-bev: Bev representation from forward-backward view transformations},
  author={Li, Zhiqi and Yu, Zhiding and Wang, Wenhai and Anandkumar, Anima and Lu, Tong and Alvarez, Jose M},
  booktitle={Proceedings of the IEEE/CVF International Conference on Computer Vision},
  pages={6919--6928},
  year={2023}
}

@inproceedings{sun2021loftr,
  title={LoFTR: Detector-free local feature matching with transformers},
  author={Sun, Jiaming and Shen, Zehong and Wang, Yuang and Bao, Hujun and Zhou, Xiaowei},
  booktitle={Proceedings of the IEEE/CVF conference on computer vision and pattern recognition},
  pages={8922--8931},
  year={2021}
}

@inproceedings{qin2022geometric,
  title={Geometric transformer for fast and robust point cloud registration},
  author={Qin, Zheng and Yu, Hao and Wang, Changjian and Guo, Yulan and Peng, Yuxing and Xu, Kai},
  booktitle={Proceedings of the IEEE/CVF conference on computer vision and pattern recognition},
  pages={11143--11152},
  year={2022}
}

@inproceedings{he2016deep,
  title={Deep residual learning for image recognition},
  author={He, Kaiming and Zhang, Xiangyu and Ren, Shaoqing and Sun, Jian},
  booktitle={Proceedings of the IEEE conference on computer vision and pattern recognition},
  pages={770--778},
  year={2016}
}
}

\appendix
\clearpage
\setcounter{page}{1}
\maketitlesupplementary

In the supplementary materials, we elaborate on the following five components to further support our paper:

\begin{enumerate}[label=\Alph*]
    \item Additional details about HOLO-CA.
    \item Ablation of loss functions.
    \item Results on the Argoverse Dataset.
    \item More detailed run time analysis.
    \item Qualitative visualization.
\end{enumerate}

\section{Additional Details about HOLO-CA}
\textbf{Structure illustration.} As shown in \cref{Fig:HOLO_CA}, given a pair of semantic features $(\mathbf{\Phi}_{bev},\mathbf{\Phi}_{map})$, the HOLO-CA module aims to estimate the relative pose through attention-based feature fusion.
We first generate patch embeddings by applying a $4\times4$ convolution with stride 4 to downsample the input images, yielding the patch-level feature maps $\mathbf{F}_{bev}$ and $\mathbf{F}_{map}$.

To model intra- and inter-feature interactions, We adopt self-attention and cross-attention respectively. Given BEV and map features $\mathbf{F}_{bev}$ and $\mathbf{F}_{map}$, 
they first undergo self-attention within each modality, followed by cross-attention between the two modalities. For an input sequence $X$, the self-attention and cross-attention output is computed as
\begin{equation}
   \mathrm{selfAttn}(X) = \mathrm{softmax}\!\left(\frac{Q_iK_i^{\top}}{\sqrt{d}}\right)V_i, 
\end{equation}
\begin{equation}
   \mathrm{crossAttn}(X) = \mathrm{softmax}\!\left(\frac{Q_iK_j^{\top}}{\sqrt{d}}\right)V_j, 
\end{equation}
where $Q$, $K$, and $V$ denote the query, key, and value matrices, respectively. $i,j$ indicate different modality.
In this way, BEV patches query map features, and map patches query BEV features, allowing the network to learn correspondences across the two modalities effectively.

Finally, the attention-enhanced features are concatenated and passed to the subsequent pose estimation layers. The pose estimation layers act as a decoder that transforms the fused features into either  corner displacements for homography estimation or a direct 3-DoF pose.

If the network regresses corner displacements, the updated four corner points parameterize a perspective transform used to warp input BEV features $\mathbf{\Phi}_{bev}$:
\[
\mathbf{\Phi}_{bev}^{(t+1)} = \mathcal{H}(\mathbf{p}^{(t+1)}) \cdot \mathbf{\Phi}_{bev}^{(t)},
\]
where $\mathcal{H}$ denotes a differentiable homography estimator.  
The warped image is re-encoded at the next iteration, enabling coarse-to-fine refinement.

\textbf{Implementation Details.} The input semantic features $\mathbf{\Phi}_{bev}$ and $\mathbf{\Phi}_{map}$ share the same spatial resolution as those fed into the HOLO, while the patch embeddings $\mathbf{F}_{bev}$ and $\mathbf{F}_{map}$ are of size $64 \times 64 \times 256$.
During feature fusion, we compress each patch token from 256 dimensions to 96 dimensions for constructing the query, key, and value vectors. We use a three-layers attention block.

\begin{table}[ht]
  \centering
  \caption{Comparison of localization accuracy across different numbers of iterations.}
  \scalebox{0.8}{
  \setlength{\tabcolsep}{3pt}{
  \begin{tabular}{ccccccccc}
    \toprule
    \multirow{2.5}*{Iteration} &
    \multicolumn{4}{c}{\centering Position Recall@$Xm$ $\uparrow$} &
    \multicolumn{4}{c}{\centering Orientation Recall@$X^\circ$ $\uparrow$} \\
    \cmidrule(r){2-5} \cmidrule(r){6-9}
    & 
    \multicolumn{1}{c}{\centering $1m$} &
    \multicolumn{1}{c}{\centering $2m$} &
    \multicolumn{1}{c}{\centering $5m$} &
    \multicolumn{1}{c}{\centering $10m$} &
    \multicolumn{1}{c}{\centering $1^\circ$} & 
    \multicolumn{1}{c}{\centering $2^\circ$} &
    \multicolumn{1}{c}{\centering $5^\circ$} &
    \multicolumn{1}{c}{\centering $10^\circ$} \\
    \midrule
    1 & $21.47$ & $46.70$ & $77.71$ & $90.02$ & $39.52$ & $76.13$ & $94.16$ & $98.41$ \\
    2 & $23.49$ & $49.60$ & $79.00$ & $90.52$ & $39.52$ & $74.27$ & $94.96$ & $99.20$ \\
    \bottomrule
  \end{tabular}}}
  \label{tab:app_holoca_itr}
\end{table}
\begin{table}[ht]
  \caption{Comparison of the model performance across different numbers of iterations.}
  \label{tab:app_holoca_perf}
  \centering
  \begin{tabular}{@{}ccc@{}}
    \toprule
    Iteration & $\text{GFLOPs} \downarrow$ & $\text{FPS} \uparrow$ \\
    \midrule
    1 & $109.10$ & $23.47$\\
    2 & $118.93$ & $16.41$\\
    \bottomrule
  \end{tabular}
\end{table}
\textbf{Additional Experiments.} We evaluate the localization accuracy and model performance of HOLO-CA with homography head under different iteration numbers. Increasing the number of iterations from 1 to 2 leads to a moderate improvement in localization accuracy. However, unlike HOLO, HOLO-CA must recompute the feature fusion module at every iteration. As shown in \cref{tab:app_holoca_perf}, this iterative recomputation leads to rapidly escalating computational cost, which in turn causes a pronounced degradation in inference speed.

\begin{table*}[ht]
  \centering
  \caption{Localization results of various combination of loss functions on nuScenes dataset.}
  \setlength{\tabcolsep}{1.2mm}{
  \begin{tabular}{ccccccccccccc} 
    \toprule
    \multirow{2.5}*{Pose Loss} &
    \multirow{2.5}*{BEV Loss} &
    \multirow{2.5}*{Map Loss} &
    \multicolumn{4}{c}{\centering Recall@$Xm$ $\uparrow$} &
    \multicolumn{4}{c}{\centering Orientation Recall@$X^\circ$ $\uparrow$} &
    \multirow{2.5}*{APE$(m)$ $\downarrow$} & 
    \multirow{2.5}*{AOE$(^\circ)$ $\downarrow$} \\
    \cmidrule(r){4-7} \cmidrule(r){8-11}
    & & &
    \multicolumn{1}{c}{\centering $1m$} &
    \multicolumn{1}{c}{\centering $2m$} &
    \multicolumn{1}{c}{\centering $5m$} &
    \multicolumn{1}{c}{\centering $10m$} &
    \multicolumn{1}{c}{\centering $1^\circ$} & 
    \multicolumn{1}{c}{\centering $2^\circ$} &
    \multicolumn{1}{c}{\centering $5^\circ$} &
    \multicolumn{1}{c}{\centering $10^\circ$} \\
    \midrule
    $\checkmark$ & $\checkmark$ & $\checkmark$ & $\textbf{36.41}$ & $\textbf{61.21}$ & $\textbf{84.10}$ & $\textbf{93.09}$ & $\textbf{73.74}$ & $\textbf{91.51}$ & $97.35$ & $99.20$ & $\textbf{3.37}$ & $\textbf{1.02}$ \\
    $\checkmark$ & $\checkmark$ & $ $ & $30.49$ & $58.70$ & $83.01$ & $92.77$ & $71.88$ & $91.51$ & $\textbf{98.14}$ & $99.20$ & $3.63$ & $1.04$ \\
    $\checkmark$ & $ $ & $ $ & $17.01$ & $40.91$ & $77.82$ & $90.72$ & $56.50$ & $85.68$ & $97.61$ & $\textbf{99.47}$ & $4.58$ & $1.31$ \\
    \bottomrule
  \end{tabular}}
  \label{tab:ablation_loss_func}
\end{table*}

\section{Ablation of Loss Functions}
To construct feature pairs with homography, we introduce BEV and map semantic losses to encourage semantic alignment between the two modalities. The detailed expression of semantic loss $\mathcal{L}_{\text{sem}}$ is 
\begin{equation}
\mathcal{L}_{\text{sem}}
= \mathcal{L}_{\text{BEV}}^{\text{sem}}+\mathcal{L}_{Map}^{\text{sem}}.
\end{equation}
As shown in \cref{tab:ablation_loss_func}, adding the BEV semantic loss yields a substantial improvement in localization accuracy, indicating that this loss helps bring BEV features closer to map features in both geometry and semantics.
With the additional map semantic loss, the accuracy is further improved, suggesting that it effectively narrows the modality gap and provides higher-quality paired features.

\section{Results on the Argoverse Dataset.} 

We further conduct experiments on the Argoverse dataset. Argoverse is also an autonomous driving dataset collected in Miami and Pittsburgh. Following the official split, we use 13,122 samples for training and 5,015 samples for validation. Similar to nuScenes, Argoverse does not provide SD map data; therefore, we obtain the corresponding SD maps from OpenStreetMap (OSM). In contrast to nuScenes, this dataset does not require any alignment.

\begin{table}[ht]
  \centering
  \caption{Localization results on argoverse dataset. * denotes results directly reported in the original paper, where the model is first trained on nuScenes and then fine-tuned on Argoverse.}
  \scalebox{0.85}{
  \setlength{\tabcolsep}{3pt}{
  \begin{tabular}{ccccccc}
    \toprule
    \multirow{2.5}*{Method} &
    \multicolumn{3}{c}{\centering Position Recall@$Xm$ $\uparrow$} &
    \multicolumn{3}{c}{\centering Orientation Recall@$X^\circ$ $\uparrow$} \\
    \cmidrule(r){2-4} \cmidrule(r){5-7}
    & 
    \multicolumn{1}{c}{\centering $1m$} &
    \multicolumn{1}{c}{\centering $2m$} &
    \multicolumn{1}{c}{\centering $5m$} &
    \multicolumn{1}{c}{\centering $1^\circ$} & 
    \multicolumn{1}{c}{\centering $2^\circ$} &
    \multicolumn{1}{c}{\centering $5^\circ$} \\
    \midrule
    MapLocNet* & $23.26$ & $47.24$ & $79.13$ & $62.35$ & $86.28$ & $96.24$ \\
    HOLO & $29.38$ & $51.89$ & $77.26$ & $65.61$ & $86.33$ & $95.86$ \\
    \bottomrule
  \end{tabular}}}
  \label{tab:app_exp_arg}
\end{table}

\cref{tab:app_exp_arg} reports the performance of our method on the Argoverse dataset. Since the results of MapLocNet are obtained by first training on nuScenes and then fine-tuning on Argoverse, whereas our model is trained from scratch directly on Argoverse, with significantly fewer training samples, our Recall@5m/5° is slightly lower. However, our method substantially outperforms MapLocNet on Recall@1m/2m, demonstrating the superior localization accuracy of our approach.

\section{More Detailed Run Time Analysis}
\begin{figure}[ht]
  \centering
   \includegraphics[width=0.97\linewidth]{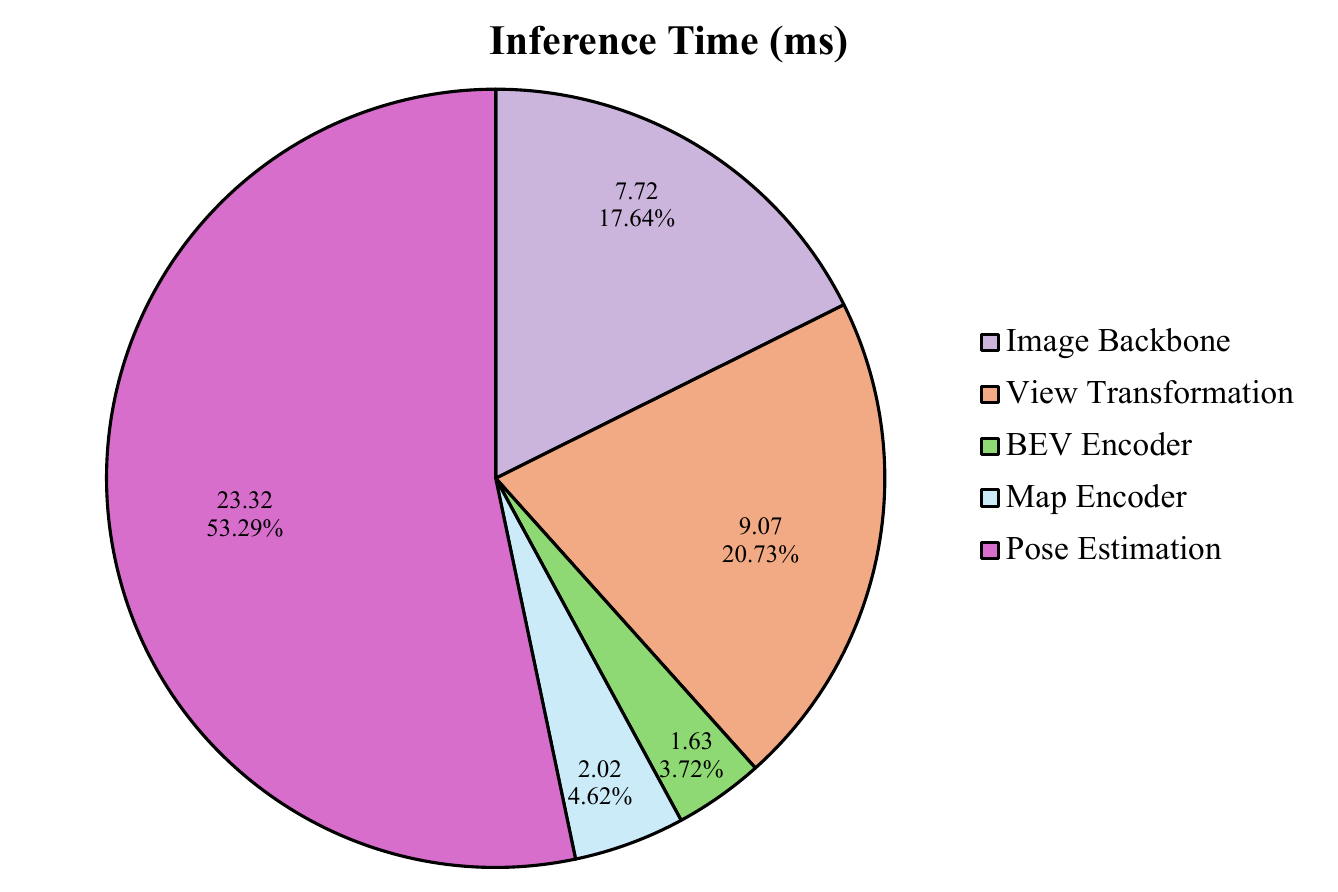}
   \caption{Detailed run time. We conducted an inference time analysis of each component of HOLO on an NVIDIA RTX A6000 GPU.}
   \label{fig:runtime}
\end{figure}
\cref{fig:runtime} presents the detailed inference-time breakdown of HOLO under the 6-iteration setting. Overall, the frontend feature perception stage and the backend pose estimation stage each account for roughly half of the total runtime, indicating a balanced computational load between the two. Within the feature perception stage, the image backbone and the BEV view transformation dominate the computation cost, representing the primary bottlenecks of the pipeline. These observations suggest that further optimization of image encoding and BEV generation will be crucial for improving the overall efficiency of the system in real-world deployment.

\section{Qualitative Visualization}

\begin{figure*}[ht]
  \centering
  \includegraphics[width=0.98\linewidth]{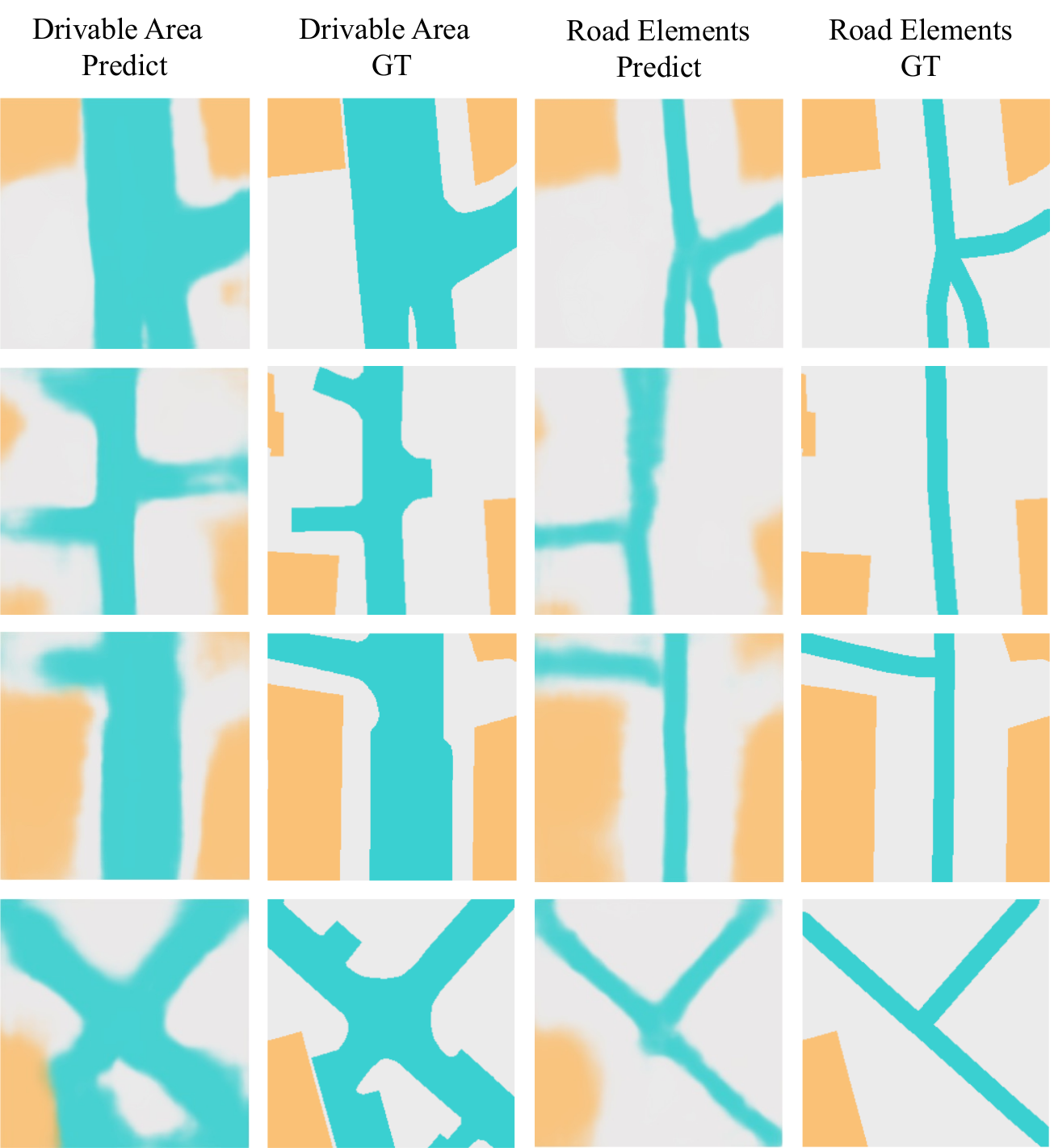}
  \caption{The segmentation results of the BEV perception module under the supervision of drivable area and road elements, respectively.}
  \label{fig:seg_results}
\end{figure*}

\textbf{Segmentation Visualization.} \cref{fig:seg_results} shows the semantic segmentation results of the BEV perception module when supervised by the drivable areas provided by nuScenes and the road elements extracted from OSM, respectively. Each row in the figure corresponds to segmentation results at the same physical location. As illustrated by the second row, the road annotations in the SD map deviate from the actual road structure observed by the camera. The drivable-area ground truth indicates the presence of a fork at that location, whereas the SD map labels fail to provide this information. Consequently, the BEV perception module’s predictions become misaligned with the SD map, degrading localization accuracy. In contrast, supervision from drivable areas offers annotations that closely match real-world observations and contain richer structural details, enabling more accurate localization.

\begin{figure*}[ht]
  \centering
  \includegraphics[width=0.98\linewidth]{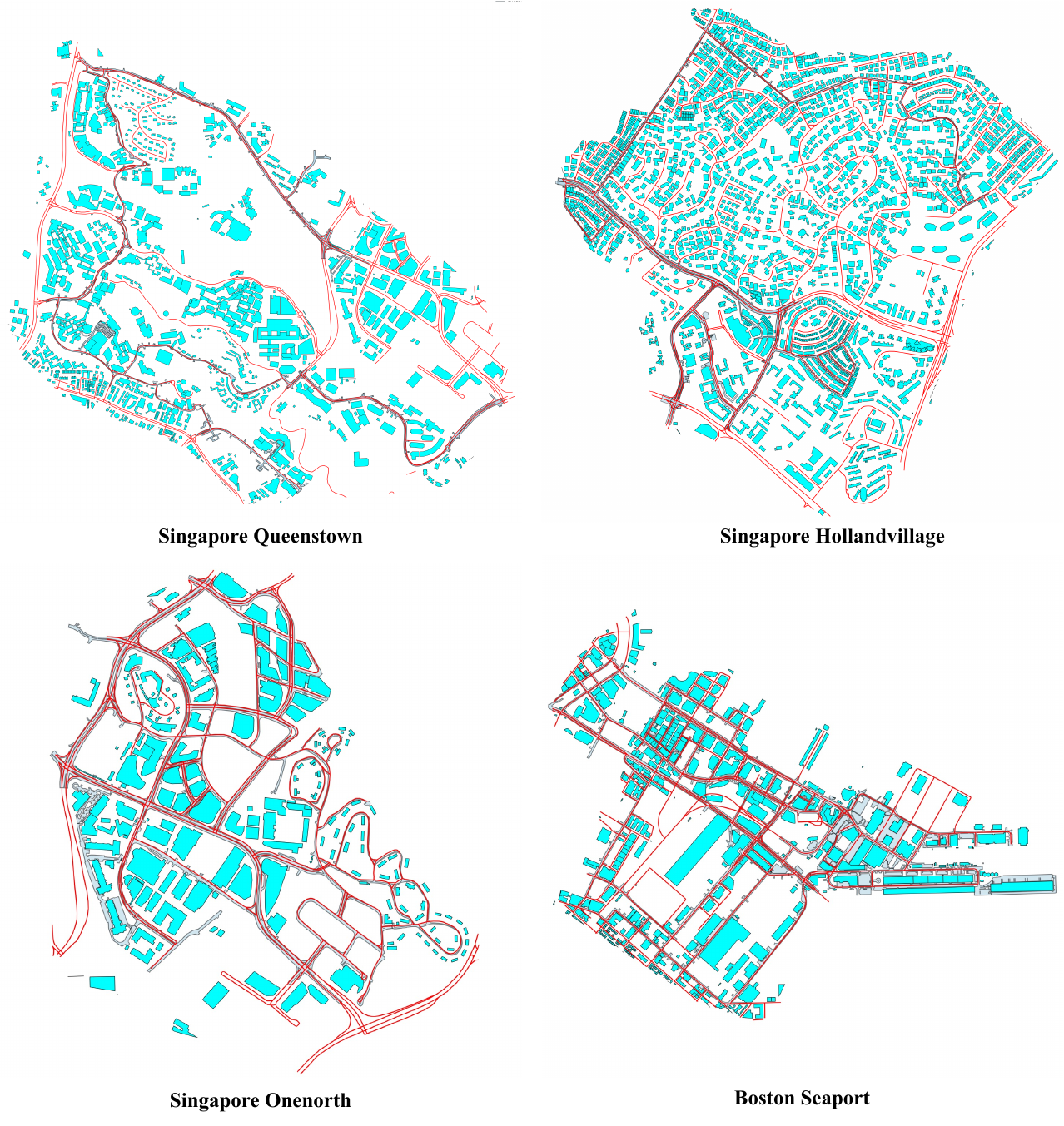}
  \caption{The alignment results between the nuScenes HD map and the OSM map collected in our study. The red curves represent the roads extracted from the SD map, while the light blue regions correspond to the drivable-area layer in the HD map. The cyan regions indicate the building areas in the SD map.}
  \label{fig:nuScene_map_align}
\end{figure*}

\begin{figure*}[ht]
  \centering
  \includegraphics[width=0.98\linewidth]{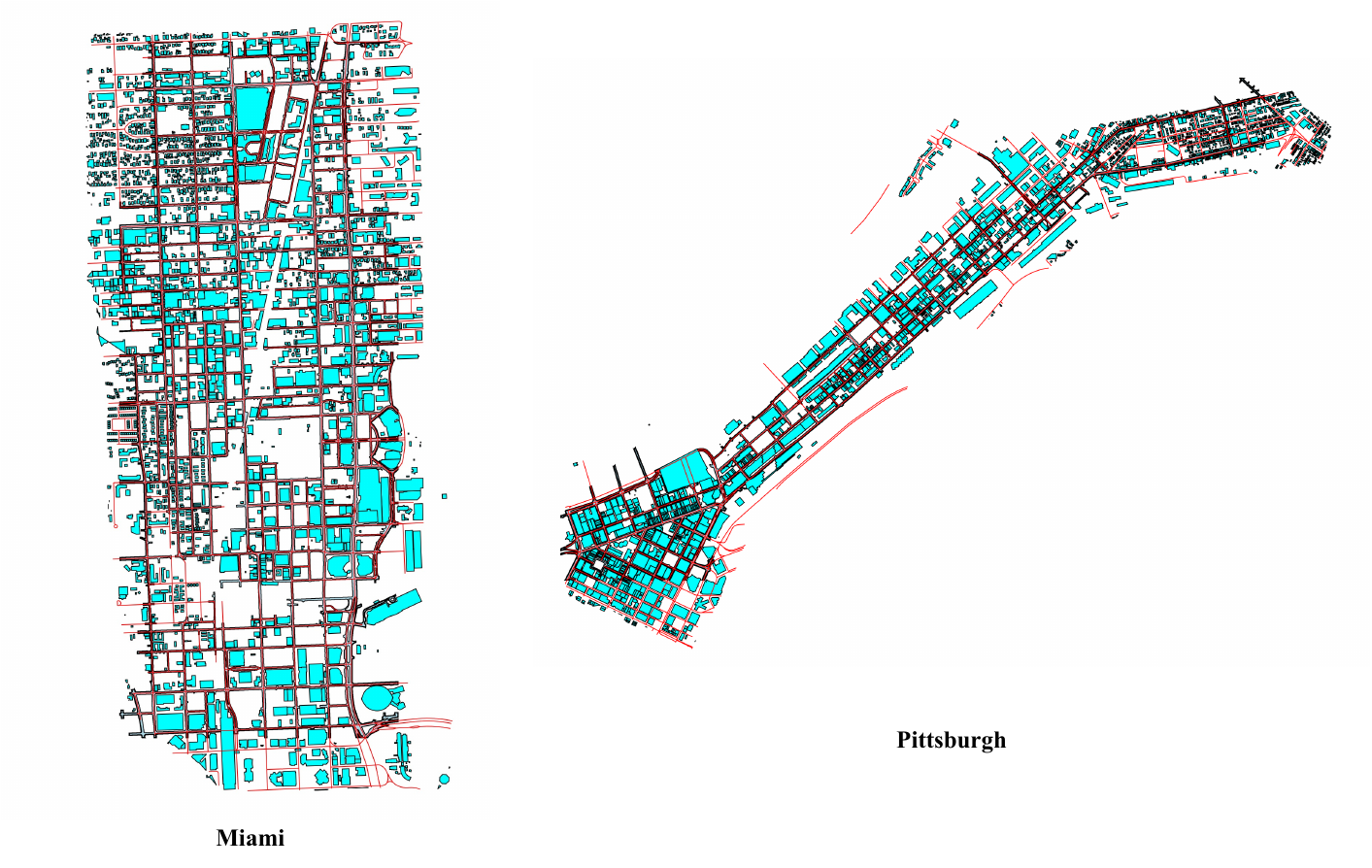}
  \caption{The alignment results between the Argoverse HD map and the OSM map collected in our study. The red curves represent the roads extracted from the SD map, while the light blue regions correspond to the drivable-area layer in the HD map. The cyan regions indicate the building areas in the SD map.}
  \label{fig:argoverse_map_align}
\end{figure*}

\textbf{Map Alignment Visualization.} Based on the procedure for converting WGS84 coordinates into the ENU coordinate system, the alignment results between our SD map and the nuScenes HD map are shown in \cref{fig:nuScene_map_align}. The road geometry from the SD map (red curves) aligns well with the drivable-area layer of the HD map (light blue regions). However, due to the limited positional accuracy of the SD map, certain misalignments still occur. In addition, because the SD map is not updated in a timely manner, some road segments that appear in the HD map are missing from the SD map. \cref{fig:argoverse_map_align} presents the alignment results between the Argoverse HD map and our collected SD map. Since the original map and the SD map share the same coordinate system, no additional alignment procedure is required.

\begin{figure*}[ht]
  \centering
  \includegraphics[width=0.97\linewidth]{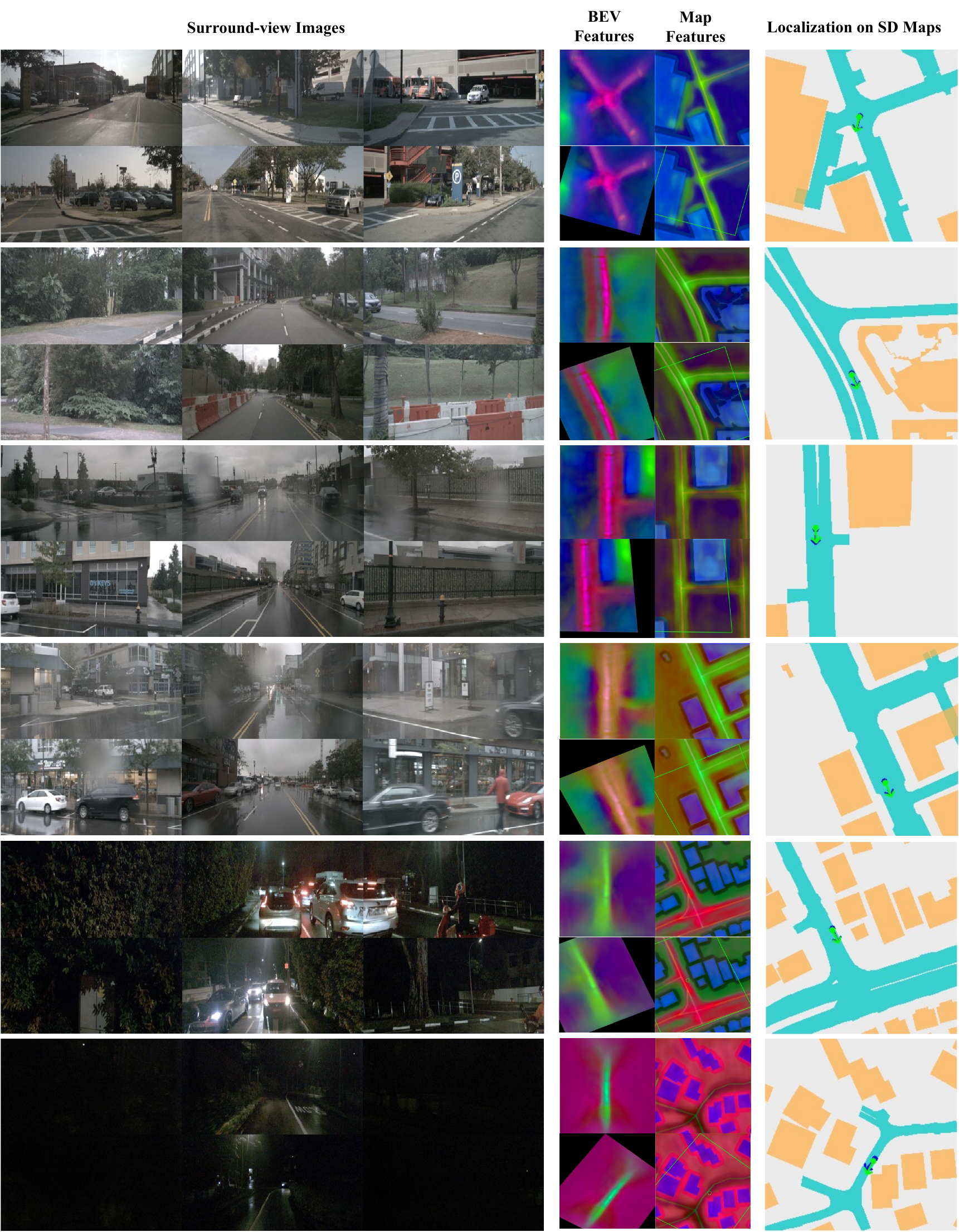}
  \caption{Additional qualitative results for visual localization on nuScene dataset.}
  \label{fig:app_nuscene_loc}
\end{figure*}

\textbf{Qualitative Results.} \cref{fig:app_nuscene_loc} further presents additional qualitative results of HOLO on the nuScenes dataset. Beyond typical daytime scenes, the fourth to sixth rows illustrate localization performance under more challenging conditions such as rainy weather and nighttime driving. These results demonstrate that our method maintains stable localization accuracy despite variations in illumination, adverse weather, and other forms of visual degradation, highlighting its strong robustness across diverse environments.

\begin{figure*}[ht]
  \centering
  \includegraphics[width=0.99\linewidth]{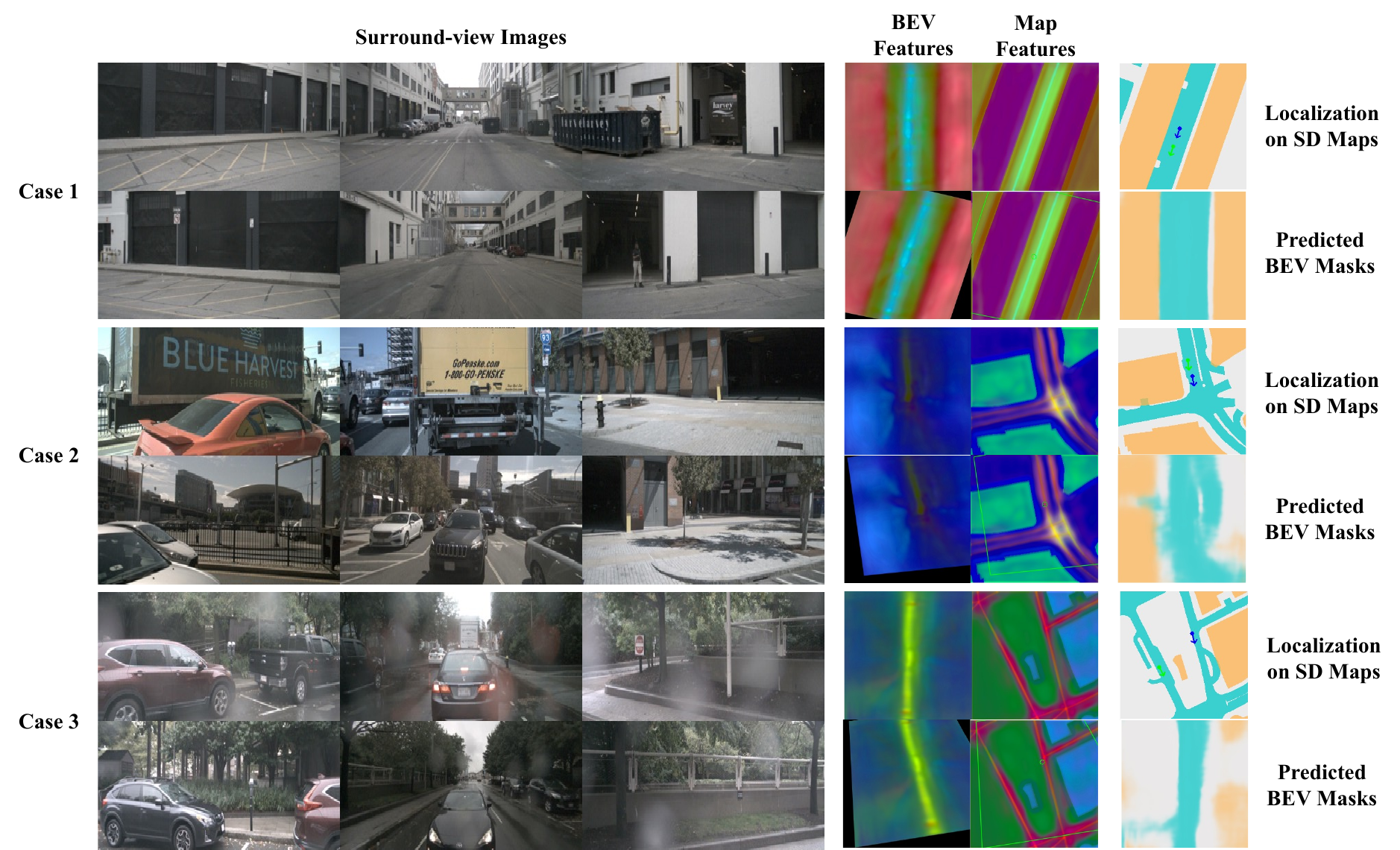}
  \caption{Visualization of three typical failure cases in our localization framework.}
  \label{fig:app_failure}
\end{figure*}

\textbf{Failure Case Study.} As shown in \cref{fig:app_failure}, we also examine several representative failure cases of the model’s localization results, which may offer insights for improving future approaches and inspire further research in this direction.

\textit{Case 1.} The first category of large localization errors primarily arises from the sparsity of discriminative features along the road direction. As shown in the visualization, both the BEV features and the map features exhibit very limited variation along the longitudinal direction of the road. This lack of distinctive cues provides insufficient geometric constraints for the model, leading to significant localization drift along the road axis. In contrast, the features vary more prominently in the direction perpendicular to the road, allowing the model to establish stronger constraints and achieve higher localization accuracy in that dimension.

\textit{Case 2.} The second source of localization error is the presence of dense traffic. Since our homography estimation relies primarily on static landmarks such as road boundaries and building contours, heavy traffic can significantly degrade the quality of BEV perception. As illustrated in the figure, densely packed vehicles obscure large portions of the roadway, resulting in blurred or incomplete road structures with weak geometric cues. This degradation makes it more difficult to estimate a reliable homography, ultimately leading to increased localization errors.

\textit{Case 3.} The third type of large localization error arises from the presence of many structurally similar regions in the SD map. Because the BEV perception has a limited field of view while the SD map typically covers a much larger area, the model may encounter multiple regions in the map that share highly similar local patterns with the observed BEV features. This often leads to ambiguous associations and incorrect matches. In the illustrated example, the local geometric structure around the ground-truth position closely resembles that of the model’s estimated position, making it difficult for the network to disambiguate between the two and ultimately resulting in a significant localization offset.

\end{document}